\documentclass[preprint]{elsarticle}

\usepackage{graphicx}
\usepackage{wasysym}
\usepackage{amsfonts}
\usepackage{amssymb}
\usepackage{xcolor}
\usepackage[T1]{fontenc}
\usepackage[a4paper, total={6in, 9in}]{geometry}

\renewenvironment{abstract}
 {\small
  \begin{center}
  \bfseries \abstractname\vspace{-.5em}\vspace{0pt}
  \end{center}
  \list{}{%
    \setlength{\leftmargin}{20mm}
    \setlength{\rightmargin}{\leftmargin}%
  }%
  \item\relax}
 {\endlist}

\begin{document}

\title{Investigating Navigation Strategies in the Morris Water Maze through Deep Reinforcement Learning}

\author{Andrew Liu}
\ead{aliu@math.utah.edu}
\author{Alla Borisyuk}
\ead{borisyuk@math.utah.edu; Dept. of Mathematics, 155 E 1400 S, Salt Lake City, UT 84109, USA}

\maketitle{}

\begin{abstract}
Navigation is a complex skill with a long history of research in animals and humans.
In this work, we simulate the Morris Water Maze in 2D to train deep reinforcement learning agents.
We perform automatic classification of navigation strategies, analyze the distribution of strategies used by artificial agents, and compare them with experimental data to show similar learning dynamics as those seen in humans and rodents.
We develop environment-specific auxiliary tasks and examine factors affecting their usefulness.
We suggest that the most beneficial tasks are potentially more biologically feasible for real agents to use.
Lastly, we explore the development of internal representations in the activations of artificial agent neural networks.
These representations resemble place cells and head-direction cells found in mouse brains, and their presence has correlation to the navigation strategies that artificial agents employ.

\end{abstract}


\section{Introduction}

Efficient navigation is essential for intelligent agents to achieve their goals in the world and has a rich history of study in animals, humans, as well as robots. 
The many approaches with which navigation can be explored make it an interesting task to simulate and train artificial agents with. 
In this paper we consider navigation learning in the deep reinforcement learning (RL) framework \cite{mnih:2016, kulkarni:2016, zhu:2016} as a model for real intelligence. 

Specifically, we focus on an environment known as the Morris Water Maze (MWM) task \cite{morris:1984}, which has been used extensively to study human and rodent navigation.
In the classical version of this task, a mouse is placed in a circular pool filled with opaque water that hides a platform.
Over consecutive trials the mouse gradually learns to use proximal and distal cues to navigate towards the platform with increasing speed, accuracy, and rate of success.

Various details about experimental protocol in the MWM can be altered to observe their effects on learning \cite{hodges:1996}.
The task can be used to explore how damage or deficiencies in the brain affect the ability to learn to navigate \cite{morris:1984}.
The task has also been mimicked in virtual reality to test human capabilities \cite{goodrich:2010, schoenfeld:2017}, and is a useful tool to study how differences in innate and environmental factors influence navigational strategy usage \cite{barhorst:2021, padilla:2017}

We develop a simulated version of the MWM for training artificial RL agents and explore factors influencing training.
Our simulation environment is computationally simpler than other navigation RL tasks that have previously been studied \cite{kempka:2016, mnih:2016}, allowing faster experimental iteration while maintaining interesting training dynamics and learned behaviors.
To our knowledge it is also the first 2D replication of the MWM in RL.
We are particularly interested in drawing comparisons between behaviors learned in humans or rodents and those learned by our artificial agents. 
To this end we train a machine learning model to automatically classify navigation trajectories and analyze the effects of the availability of different global cues on learned behaviors.
We identify five behavior types within our agents - `stuck', `circling', `corner testing', `indirect navigation', and `direct navigation'.
The last three of these are considered spatial navigation strategies (as opposed to non-spatial ones).

Within our MWM environment, we develop several training conditions that provide varying amounts and types of global landmarks for the agents to navigate by. 
For example, in one of the more difficult variations of the MWM that we focus on, the only available landmark is a small poster. 
We find that different training conditions lead to the development of distinct navigation strategy preferences. 
Individual agents also exhibit a variability of behaviors across episodes. 
On average and in most conditions, during early stages of training non-spatial strategies like searching and exploring are used, and as training continues spatial navigation becomes more common.
This trend in learning dynamics is similar to those reported in rodents and humans \cite{schoenfeld:2010}.

To improve training effectiveness and influence the development of strategy preferences, we explore the approach of introducing auxiliary tasks.
Auxiliary tasks \cite{jaderberg:2016, kartal:2019, mirowski:2016}, where agents are assigned tasks alongside the main RL goal, have been applied to improve learning.
Auxiliary tasks are learned by optimizing the same weights used by the agent's policy network, and encourage agents to learn additional information about the environment.
Past work on auxiliary tasks have often focused on specific tasks designed to improve learning rates in specific RL settings.
In contrast, we explore and compare a range of auxiliary tasks across different task classes.
We find that in the MWM environment, tasks encouraging exploration can improve learning rate early in training, and a range of categorical supervised auxiliary tasks improve the frequency and consistency of spatial strategy development.
Hence, these latter tasks help agents converge to more performant final policies after training.
We suggest that the tasks that provide the greatest benefit to our RL agents are those that would be more feasible for humans or rodents to implement in real navigation learning.

Finally, we measure the activity of the units in agents' networks across navigation trajectories to examine the agents’ ``representations'' of the environment. 
In particular, we explore the development of spatial-location-specific or direction-sensitive representations, which are similar to spatial activity maps observed in hippocampal place cells or head-direction cells respectively. 
We find that an increased presence of direction-sensitive representations (and to a lesser extent, location-sensitive representations as well) correlates with increased direct navigation strategy usage in agents, and increased MWM performance accordingly.
We also characterize the changes induced in these ``neuronal'' representations by the assignment of auxiliary tasks, in particular finding that the tasks we would expect to benefit from location or direction knowledge encourage development of the respective representations.

\section{Materials and Methods}

\subsection{Reinforcement Learning}
\label{sec:reinforcement_learning}

We apply the reinforcement learning framework \cite{sutton:2018} where an agent interacts with an environment in discrete time steps to maximize rewards earned.
We treat the environment as a Partially Observable Markov Decision Process (POMDP) defined by the tuple $(S, A, P, R, \Omega, O)$.
At each time step $t$, the environment is in state $s_t \in S$.
An observation $o_t \in \Omega$ which provides some partial information about the state is given to the agent, defined by the mapping $O: S \rightarrow \Omega$.
Given the observation $o_t$, the agent performs an action $a_t \in A$, which affects the state according to the transition function $P: S \times A \rightarrow S$, and the agent receives reward $r_t$ given by the reward function $R: S \times A \rightarrow \mathbb{R}$.
More generally the functions $P, R, O$ may map to probability distributions, but in our environment they are deterministic functions.
To learn to operate in a POMDP the agent's neural network is given a recurrent layer, allowing it to have memory or a hidden state $h_t \in \mathbb{R}^k$ where $k$ is the number of nodes in the recurrent layer.

The agent's goal is to learn a policy $\pi(a_t | o_t,h_t) = \mathbb{P}[a = a_t | o = o_t, h_t=h]$ which outputs actions at each time step to maximize rewards.
$\gamma \in [0, 1)$ is the discount factor and the sum of discounted rewards starting from time $t$ is given by 
\begin{equation}
G_t = \sum_{k=0}^\infty \gamma^k r_{t+k+1}
\end{equation}
which is also known as the return.
The policy will be parameterized by $\theta$, which in our case contains the neural network parameters which are used to generate $\pi_\theta$.
The agent will also learn to approximate the value of the current observation, which is the expected return given that the agent follows its policy
\begin{equation}
V^{\pi_\theta}(o, a, h) = \mathbb{E}^{\pi_\theta}[G_t | o_t=o, a_t=a, h_t=h].
\end{equation}

In the context of a POMDP, we also think about \textit{representations}.
Formally, a representation is a function which maps observations and hidden state to d-dimesional features $\phi: O \times \mathbb{R}^k \rightarrow \mathbb{R}^d$.
We think of useful representations as features that allow the agent to keep track of information about the environment state. 
These representations or features are then used in downstream computation for example of $V$ or $\pi$.

\subsection{2D Simulated Navigation Environment}
\label{sec:mwm_env_descriptions}

\begin{figure}
\centering
\includegraphics{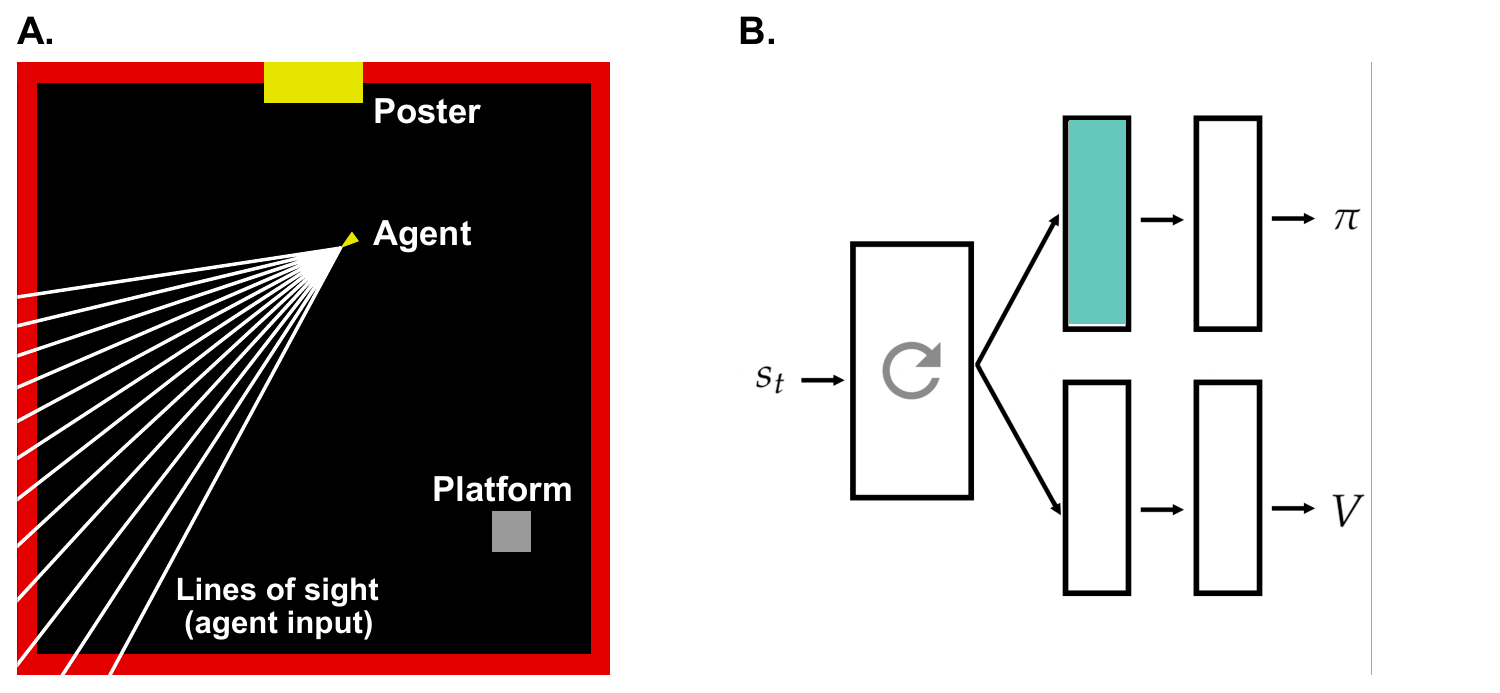}
\caption{\textbf{A.} A graphical representation of the simulated 2D Morris Water Maze task environment. 
The yellow triangle corresponds to the agent with white vision lines extending out.
Each vision line returns the color and distance of the wall/poster that it intersects with. 
There is a visible yellow poster on the north wall, and the box in the south-east corner represents an invisible platform that is the agent's goal to navigate to.
\textbf{B.} A depiction of the RL actor-critic neural network.
Each box represents a fully-connected feed-forward layer in the neural network, and the left-most layer is a gated recurrent unit layer.
$\pi$ and $V$ represent the policy and value outputs, respectively.
The teal-colored box shows the network layer used to measure developed agent representations.}
\label{fig:mwm_poster_env_nn}
\end{figure}

\begin{figure}
	\centering
	\includegraphics{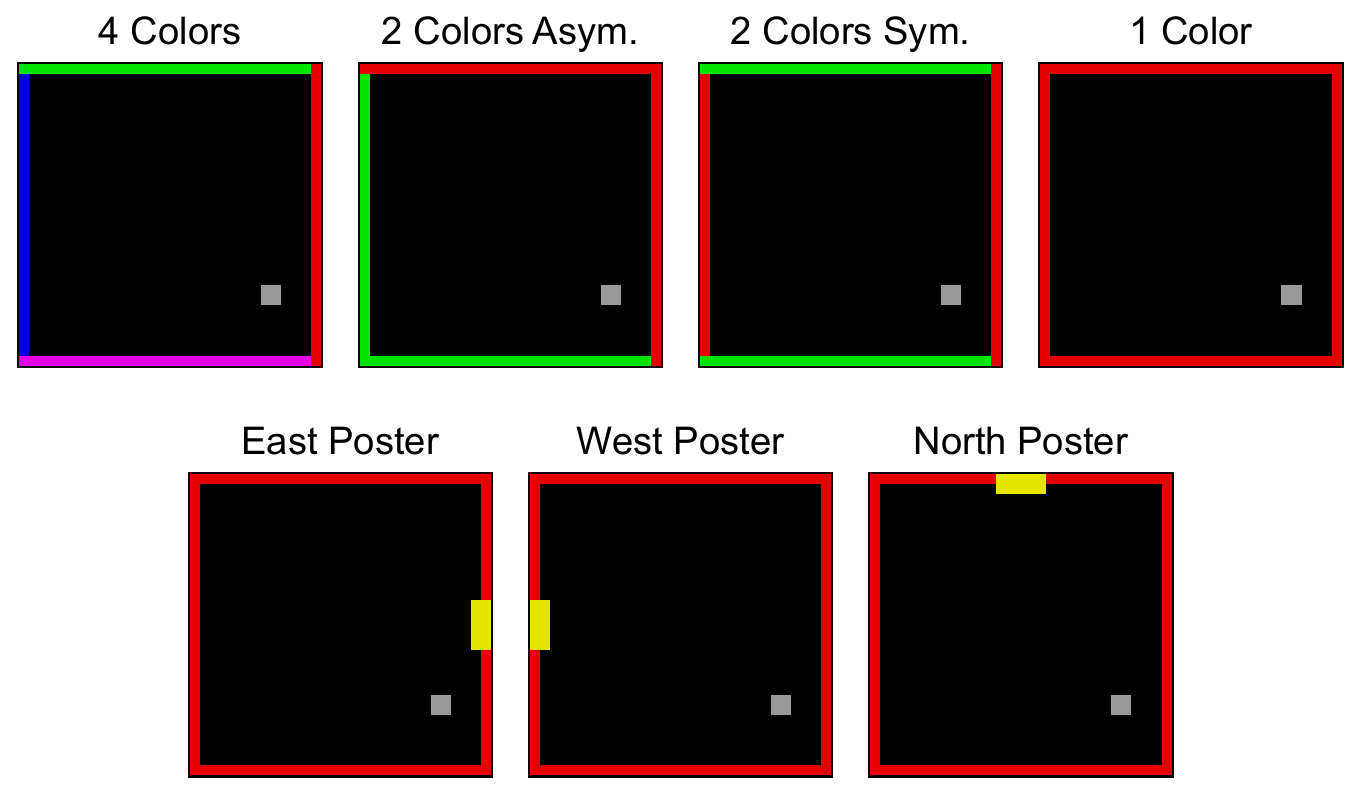}
	\caption{Variations of the simulated MWM task. \textbf{Top:} No posters are given, but different colors of walls are used as navigation cues. \textbf{Bottom:} All walls have the same color, but a small poster is given as a navigation cue.}
	\label{fig:mwm_env_types}
\end{figure}

To conduct navigation experiments, we create a simulated 2D navigation environment that is run in Python.
In this environment, the RL agent has $12$ sight lines uniformly distributed across a fixed $1$ radian field of view extending from the head of the agent, centered in the direction the agent is facing (Fig \ref{fig:mwm_poster_env_nn}A).
Each sight line returns both the color (with unique colors represented by arbitrary numbers) and distance to the intersecting wall, giving the agent an observation $o_t \in \mathbb{R}^{24}$ at each time step.
The agent is allowed to pick from four possible actions: a left or right turn of fixed degree, a forward movement of fixed speed, or no action.

Fig \ref{fig:mwm_poster_env_nn}A shows an example of the environment that the agent experiences.
The navigation space is a box with size $[0, 300] \times [0, 300]$ with non-dimensional units, and the target goal is an invisible square with a side length of $20$ units (shown as a gray box in Fig \ref{fig:mwm_poster_env_nn}).
Each forward action moves the agent a fixed $10$ units per time step, and each turning action rotates the agent's faced angle by $0.2$ radians per time step.
This is an episodic RL task \cite{sutton:2018}, where an episode ends and the simulation is reset when the agent reaches the platform, or after $200$ time steps, whichever comes first.
At each time step, the agent is given a reward of $0$, and a reward of $1$ when the platform is reached.
At the start of each episode, the agent's position and angle are set uniformly at random, with a minimum distance of $50$ units from the center of the goal and $30$ units from any wall.

We train RL agents in several variations of the simulated MWM task that provide different amounts and types of global cue for navigation.
These variations are shown in Fig \ref{fig:mwm_env_types}, and for the majority of the paper we study agents trained in the `North Poster' variation.

Previous examples of deep reinforcement learning in navigation \cite{jaderberg:2016, kulkarni:2016, mirowski:2016, zhu:2016} often performed experiments in 3D environments, such as Labyrinth \cite{mnih:2016} or ViZDoom \cite{kempka:2016}.
In these 3D environments, agents are typically given a two-dimensional array of RGB pixel values as visual input. 
The neural networks used in these cases include convolution layers to handle pixel inputs, increasing learning complexity due to their numerous learnable parameters.

We opt for a simpler 2D environment for ease of simulation and faster training time.
Our agents use comparatively shallow and easy-to-train neural networks, while still demonstrating a range of interesting behaviors and learning dynamics.

\subsection{Agent Network and Training Algorithm}
\label{sec:network_training}

Fig \ref{fig:mwm_poster_env_nn}B depicts the neural network architecture that the agent is trained with.
Observations are first fed into a shared gated recurrent unit (GRU) \cite{cho:2014}, which is a type of recurrent network layer. 
The GRU output is then passed to two parallel sets of two fully connected layers, which output either the policy $\pi$ or the value estimate of the current state $V$.
This is known as an actor-critic network, where the actor ($\pi$) decides actions and the critic evaluates the utility of the actions taken ($V$).
Each network layer has $16$ hidden units.
We adapt an implementation of the widely-used policy gradient method, proximal policy optimization (PPO) \cite{pytorchrl:2018, schulman:2017}.
PPO has been demonstrated to robustly train agents across a range of RL environments.

The agent learns by first performing its policy in the environment to generate a batch of training examples.
Following PPO, the batch of time steps experienced are used to generate a gradient to update the agent's neural network weights.
The process then repeats, collecting each batch with each updated neural network until training completes.
In order to minimize correlations of training samples within a batch, we generate experiences using 100 parallel copies of the agent and environment.
In this paper, a single ``trial'' refers to training a naive agent from start to finish.
An ``episode'' refers to a single simulated experience in the environment, starting at time step 0 and lasting until the agent reaches the platform or time step 200 elapses, whichever comes first.
All agents are trained for a total of 3e6 time steps.

When assigning auxiliary tasks, the agent's neural network may be required to construct additional outputs.
These are generated from the actor branch of the network.
We also considered generating outputs from the critic side of the network, which did not significantly alter the performance.
We design auxiliary tasks that can require either numerical or categorical predictions from the agent.
If the output required is numerical then the prediction is computed as a linear weighted sum of the final actor layer outputs.
If the output required is categorical with $n$ possible categories, then $n$ linear outputs are generated as independent weighted sums of the final actor layer outputs, and a softmax function is performed on these $n$ linear outputs to turn them into probabilities.

\subsection{Behavior Classification}
\label{sec:behavior_methods}

\begin{figure}
\centering
\includegraphics{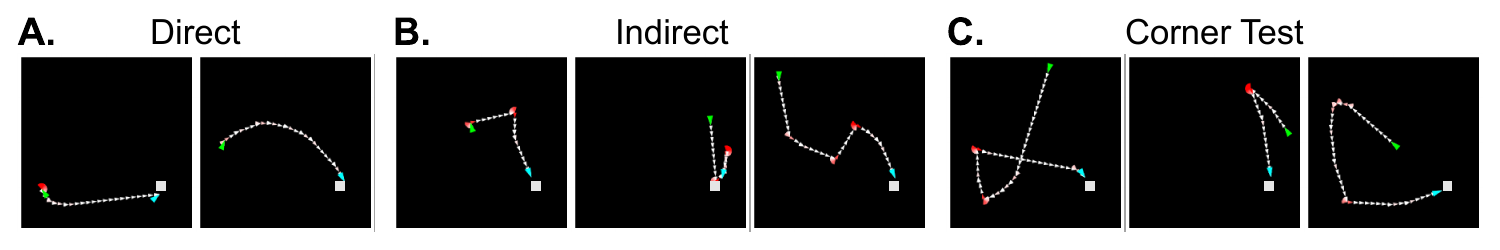}
\includegraphics{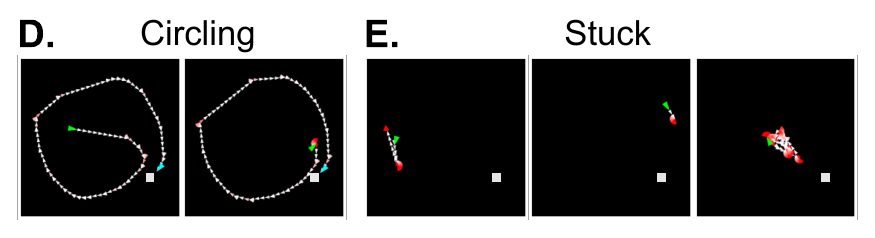}
\caption{Examples of each type of behavior that we identified. 
Each plot shows the trajectory of an agent across a single episode.
Triangles represent where the agent was at each time step in the episode.
A green triangle represents the start point of an episode, and cyan (if it exists) represents when the agent successfully navigated to the hidden platform in the bottom-right of the maze.
Triangles are colored progressively from white to red as an agent spends more time in a single location without moving forward.
\textbf{A.} Direct;
\textbf{B.} Uncertain Direct;
\textbf{C.} Corner Test;
\textbf{D.} Circling;
\textbf{E.} Stuck.}
\label{fig:trajectory_class_examples}
\end{figure}

We identify five unique types of navigation strategies as shown in Fig \ref{fig:trajectory_class_examples}.
To perform behavior classification, each agent trajectory is converted into a $224 \times 224$ pixel image like the ones shown, and a classifier model takes these images as input for training and classification.
The classifier is a pre-trained visual CNN model known as ConvNeXt \cite{liu:2022}, which is then fine-tuned to classify trajectory images.
Specifically, we use the `convnext\_tiny\_in22k' pre-trained model.
To train the classifier, we first manually label 100-200 examples of each navigation behavior type, drawn from agents trained on the North Poster MWM scenario and with various auxiliary tasks.
We perform fine-tuning using the fastai library \cite{howard:2020}, achieving around $80\%$ accuracy on set of validation examples.

This classifier is not perfectly accurate.
One challenge is that some trajectories can arguably fall into multiple categories of classification.
For example, a trajectory where an agent performed a `corner test' strategy (moving to the nearest corner to guess the platform's location) and reaches the platform on the first try will likely be identified as a `direct' navigation episode.
However, when distinguishing between spatial strategies (direct, indirect, and corner testing) and non-spatial strategies (stuck and circling) the classifier achieves over $90\%$ accuracy.
Most classification errors occur between types of spatial strategies.
Despite minor inaccuracies, when averaging across populations of agents and multiple episodes, this model provides a general idea of the distribution of strategies employed by agents.

\subsection{Auxiliary Gradient Cosine Similarity}
\label{sec:methods_aux_gradients}

To compute the cosine similarity between RL gradients and gradients induced by supervised auxiliary tasks, we perform the following steps.
We take an agent frozen at a certain checkpoint of training and randomly initialize the MWM environment with 100 parallel copies, as done in training.
The agent first ``warms up'' the environment by running its policy for 5,000 time steps, to reduce any episode start correlations between the parallel environments.
The agent collects 20 batches of 1,600 time steps of experience each, consistent with training conditions.
It then collects 3 reference batches of 25,600 time steps each.
We consider these larger reference batches to have nearly optimal, low noise gradients.
Cosine similarity is calculated between auxiliary gradients in the 1,600-step batches and RL gradients in the 25,600-step reference batches.
We define cosine similarity of a given batch as the mean similarity between $\nabla_\theta \mathcal{L}_\mathrm{aux}$ of the batch and $\nabla_\theta \mathcal{L}_\mathrm{RL}$ for the 3 reference batches. 

Similar to agent behavior analysis, we collect data and compute gradient cosine similarities for agents across eight checkpoints in training. 
Four in early training and four in middle-to-late training.

Although these gradient cosine similarities are quite natural to define with supervised auxiliary tasks, determining how to do so with reward-based auxiliary tasks is less clear.
We use the following method: first, we collect batches as we did in the supervised case (20 batches of 1,600 steps).
We compute the standard gradient $\nabla_\theta \mathcal{L}_\mathrm{RL}$ with these batches of experiences.
Next, we remove rewards from reaching the goal in each batch, and a new gradient $\nabla_\theta \mathcal{L}_\mathrm{bonus}$ is computed with this modified batch, which we call the ``pure bonus'' gradient.
For each individual batch, cosine similarity is computed between $\nabla_\theta \mathcal{L}_\mathrm{RL}$ and $\nabla_\theta \mathcal{L}_\mathrm{bonus}$.
Note that we do not collect reference batches or compare gradients across different batches.
We only compare the RL gradient of one batch to its own pure bonus gradient.

\subsection{Representation Maps}
\label{sec:methods_representation_maps}

In Section \ref{sec:representation_analysis} we explore representations developed within the neural networks of our RL agents while performing the MWM task.
We treat the activations of individual nodes in the neural network as being components of a feature vector.
Specifically, node activations in the first fully-connected layer on the policy branch of the agent neural network are measured, visualized by a teal box in Fig \ref{fig:mwm_poster_env_nn}B.
Activations are measured during natural execution of the agents' policies.
We generate 100 randomized initial positions and directions, which are saved to be used for all representation data collection.
These are supplemented with 116 initial positions that line the outer edge of the MWM play area, and each is paired with an initial direction facing the center of the area.
The 116 positions are generated by taking 30 equidistant points along each of the four outer walls of the area and removing duplicates.
An agent performs its policy until episode completion for each of these 216 starting conditions and we save recordings of all network node activations.

To generate a spatial activation heatmap for a node, we start by dividing the $[0, 300] \times [0, 300]$ MWM area into a uniform grid of $30 \times 30$ points.
Each grid point is assigned a value that is a weighted average of every activation from every time step collected in the 216 episodes for that node.
The weight is exponentially-weighted based on the distance between the position experienced at a time step and the grid point.
Specifically, for a grid point $x \in \mathbb{R}^2$ and experience position $y \in \mathbb{R}^2$, the weight is computed as 
\begin{equation}
g(x, y) = \exp(-d(x,y) / \sigma)
\label{eq:gaussian_dist}
\end{equation}
where $d(x, y)$ is the Euclidean distance between the points and we fix the parameter $\sigma$ to be $20$. 
The spatially weighted mean activation of a grid point $i$ is then computed as
\begin{equation}
a_i = \frac{1}{N}\sum_{j \in N} z_j g(x_i, y_j)  
\end{equation}
where $z_j$ is the activation of the node at time step $j$, $y_j$ is the agent's position, and $N$ is the total number of time steps collected in the 216 episodes.
Finally, we subtract each spatially weighted mean activation to get
\begin{equation}
\label{eq:spatial_heatmap}
\tilde{a}_i = a_i - \frac{1}{900} \sum_{k=1}^{900} a_k
\end{equation}
to calculate how much more or less active a node tends to be than average at points in space.
$\tilde{a}_i$'s form the spatial representation heatmap for a single node.

The process for generating direction maps for a node is analogous.
Angles of the unit circle $[-\pi, \pi]$ are uniformly divided into 100 grid points.
Once again, each angle grid point is assigned a value that averages activations of the node, weighted by distances between the angle the agent faced at each time step and the grid angle.
Equations (\ref{eq:gaussian_dist})-(\ref{eq:spatial_heatmap}) are identical, and $\tilde{a}_i$'s form direction heatmaps calculated in this manner.
\\

\section{Results}

\subsection{Navigation Learning in the Morris Water Maze}

\begin{figure}[t]
\centering
\includegraphics{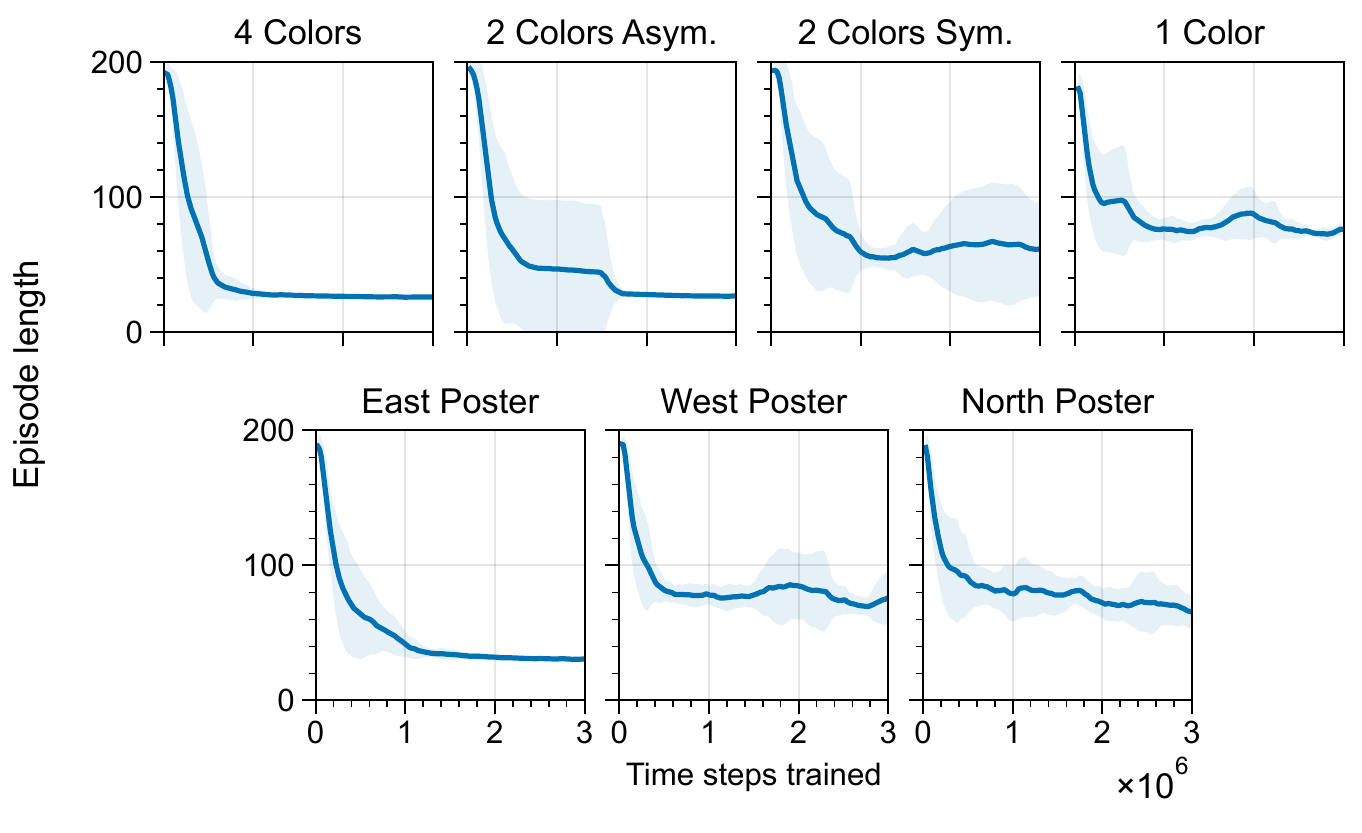}
\caption{Learning curves for agents trained in variations of the MWM environment. Solid lines show the mean performances of $10$ agents and shaded areas show $\pm 1$ standard deviation in performance.}
\label{fig:envtype_learning_curves}
\end{figure}

We train agents in 7 variations of the 2D simulated MWM task as described in Section \ref{sec:mwm_env_descriptions} and illustrated in Fig \ref{fig:mwm_env_types}.
Learning performance can be visualized using learning curves, where performance on the RL task is shown over the course of training.
In our environments, we measure performance by episode length, where shorter episodes indicate faster navigation to the goal.
Episode lengths can have a maximum value of $200$, signifying that the agent has either not reached the platform and the episode is reset, or reached the platform on the last possible time step.

Fig \ref{fig:envtype_learning_curves} presents learning curves for each MWM variation. 
These show both the mean and standard deviation of performance across 10 agents each. 
We observe that the 4 Wall Colors, 2 Asymmetrical Wall Colors, and East Poster scenarios are the easiest for agents to learn. 
With sufficient training, they consistently converge to optimal navigation policies.
When global navigation cues are always in sight (4 and 2 Wall Colors), the agents can easily compute the position of the platform in every trial.
In the East Poster scenario, the agent may need to first turn to bring the poster into sight, but from many random initial positions, it can keep the poster constantly in sight while heading to the goal due to the poster's proximity to the goal.

The North and West Poster scenarios on the other hand are much harder to learn.
Since the agent can either turn or move forward at each time step, unnecessary turns incur a cost. 
The most optimal strategies are thus complex, involving searching for the poster, calculating current position, and navigating to the goal without turning to see the poster again.
To study this challenge, we primarily focus on the North Poster variation for most of this work. 

Lastly we highlight the 2 Symmetric Wall Colors, and 1 Wall Color MWM tasks.
In these tasks, the agent cannot uniquely decode its current position due to the symmetries of the available global navigation information.
The agent receives visual information about the distance to walls in front of it, so optimal strategies would likely involve navigating to each corner that the platform may be at, essentially performing a guess and check.

\subsection{Behavior Analysis}
\label{sec:behavior_analysis}

In addition to analyzing learning rates of our RL agents, we investigate the diversity of behaviors they exhibit while executing their policies.
We are particularly interested in understanding how various modifications to the agents' training and environment affect the distribution of strategies employed both during and after training.
Previous research in MWMs with mice and humans has utilized various algorithms to perform automatic classification of navigation behaviors \cite{gehring:2015, schoenfeld:2010}. 
Research has also explored how factors such as traumatic brain injury \cite{brody:2006} or life experience \cite{barhorst:2021, padilla:2017} contribute to strategies used.

\subsubsection{Automatic strategy classification}

For simplicity, we train a neural network specialized in visual tasks to classify episode trajectories into predetermined classes, rather than hand-crafting features of importance for classification.
Fig \ref{fig:trajectory_class_examples} shows the five different classes that we identify and consider in our agents, and Section \ref{sec:behavior_methods} desribes how the model is trained.
The `direct', `indirect', and `corner test' strategies are what we consider to be methods that employ spatial understanding. 
Direct routes are those that take few detours and generally navigate directly to the platform, while indirect routes typically head to the platform after only a few detours.

The corner test strategy is an interesting and common one that is employed in complex MWM variations. 
In this strategy, agents navigate to a corner of the maze - typically the one in the most direct line of sight upon episode initialization - before then correcting their direction and navigating to the platform.
Since the agent cannot turn and move forward simultaneously, turning may be treated as incurring a penalty by taking additional time to perform.
Instead of turning to find the platform, the agent may choose to move forward to guess where the platform is.
A portion of episodes where this strategy is employed will reward the agent with a fast episode if it happens to be facing the right direction on initialization, hence reinforcing the use of this strategy.
We still categorize this as a spatial strategy, as many agents still demonstrate direct navigation after an initial failed corner test.

Finally, we consider `circling' and `stuck' trajectories to be non-spatial navigation strategies.
Naive, untrained agents often exhibit some form of stuck behavior before much training, where they fail to reach the platform and barely move from starting locations.
Circling strategies, sometimes called `thigmotaxis' in classic MWM studies, involve the agent circling the arena on a consistent track that has high probability of eventually running into the platform. 
This behavior demonstrates that the agent has learned that the platform is a set distance away from walls.

\subsubsection{Difficulty affects strategy usage}

\begin{figure}
\centering
\includegraphics{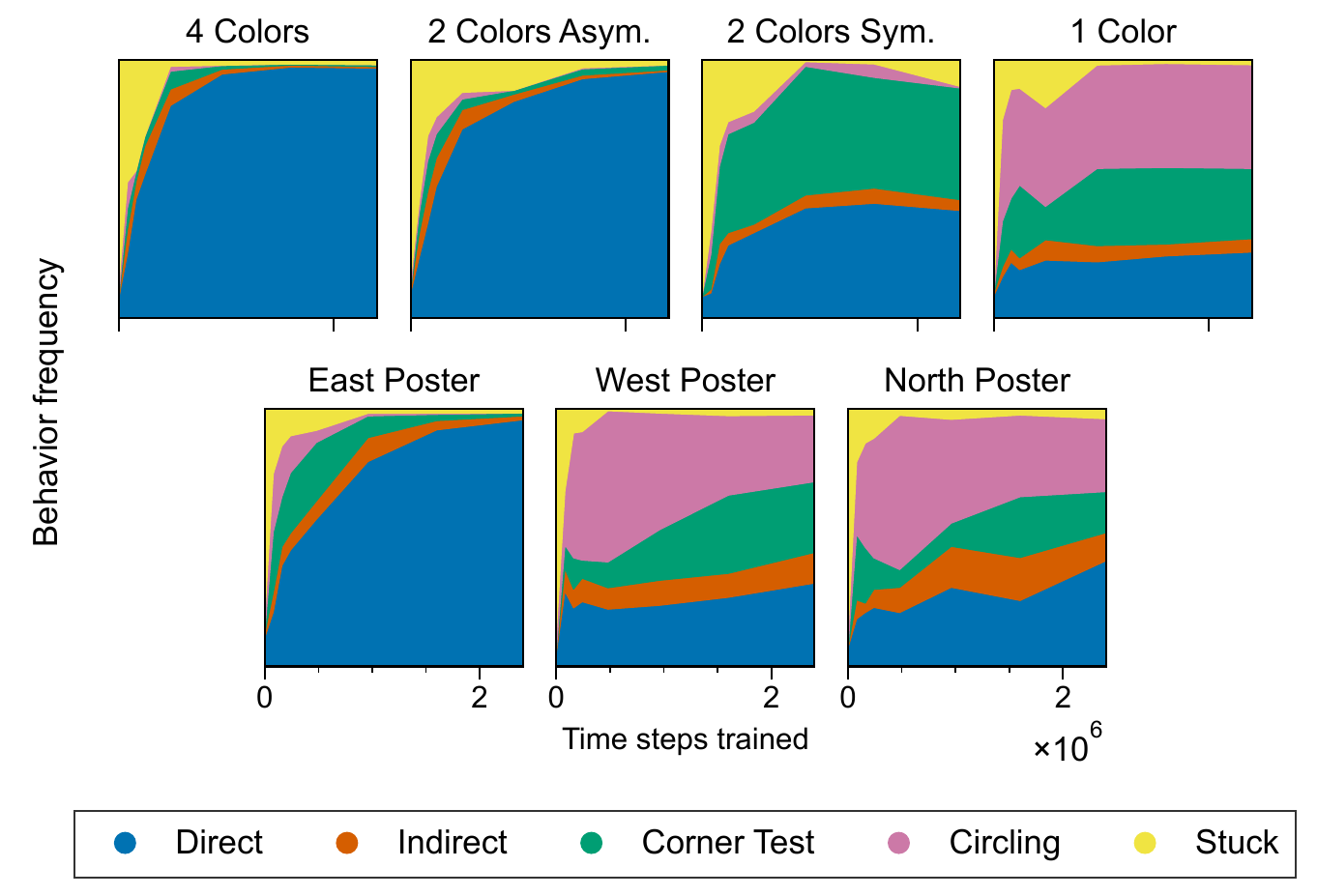}
\caption{Strategy usage across training for agents trained in each variation of the MWM task. Each plot shows strategy classification results of 10 agents each collecting 100 episodes at eight checkpoints throughout training.}
\label{fig:envtype_behaviors}
\end{figure}

We first apply our strategy classification model by classifying the behaviors of agents trained on our MWM variations.
We select eight checkpoints throughout training to evaluate agent performances, where a checkpoint is a copy of the agent's neural network weights saved at a particular point in training.
Four checkpoints are selected early in training, after approximately 0, 1e5, 2e5, and 3e5 time steps of training, as agent behaviors change most rapidly early in training.
The other four checkpoints are selected later in training, after roughly 5e5, 1e6, 1.5e6, and 2.5e6 time steps of training.

We generate 100 random initial conditions to collect episode trajectories with. 
The same initial conditions are used for every checkpoint of every agent.
We then perform automatic classification of these trajectories.
The overall usage of strategies by agents are shown in Fig \ref{fig:envtype_behaviors}.

As suggested by the learning curves previously shown in Fig \ref{fig:envtype_learning_curves}, the 4 Wall Colors, 2 Asymmetric Wall Colors, and East Poster scenarios are easy enough for the agents to almost universally learn to navigate directly to the platform.
However, we now see that in the East Poster agents, more corner testing strategies are employed compared to 4 and 2 Wall Color scenarios prior to late policy convergence.

The 2 Symmetric Wall Colors and 1 Wall Color agents have predictably consistent strategy distributions. In the 2 Wall Color case, the optimal strategy is to test the corner nearest to being faced, and then try the other corner if this fails. 
This is due to symmetries of the environment; from the agent's perspective the two corners that the platform may exist in are indistinguishable.
In $50\%$ of episodes, the first corner they test will be the correct one, and our classifier identifies these as direct navigation. 
The other $50\%$ of episodes are labeled as corner tests.
Similarly, for the 1 Wall Color case, the agent's only viable strategy is to circle around and test each corner for the platform.
We see a $25-25-50$ distribution of direct-corner test-circling classifications emerges following the random distribution of corners that are needed to be tested (either 1, 2 or 3-4 corners respectively) before the goal is found.
The classification also shows us why performance decays in the 2 Symmetric Wall Color learning curve from Fig \ref{fig:envtype_learning_curves} late in training.
Some agents deteriorate and become fully stuck, losing the ability to consistently perform effective navigation strategies.

Finally, both the West Poster and North Poster MWM variations have similar performance, with a mix of strategy usage across training. 
Circling strategies are common early on and are replaced primarily by direct and corner test strategies over time.
These developments mirror those seen in rodents, where non-spatial random or search strategies are used early in training, and more sophisticated spatial strategies are used after experience is gained \cite{schoenfeld:2017, vouros:2018}.

The results from this strategy classification align fairly well with what one might intuitively expect from each MWM variation, and this exploration confirms that the model classifies trajectories with good accuracy.
In the following sections, we focus on the North Poster case as agents here show a variety of interesting strategy usage across training.
Notably, the behavior analysis suggests that some portion of agents may find early success in reaching the platform through circling policies, but become reliant on this strategy leading to sub-optimal performance.
In the sections that follow, we are motivated to look for interventions in training methods to improve learning efficiency and increase spatial strategy usage.

\subsection{Training Batch Size}

\begin{figure}
\centering
\includegraphics{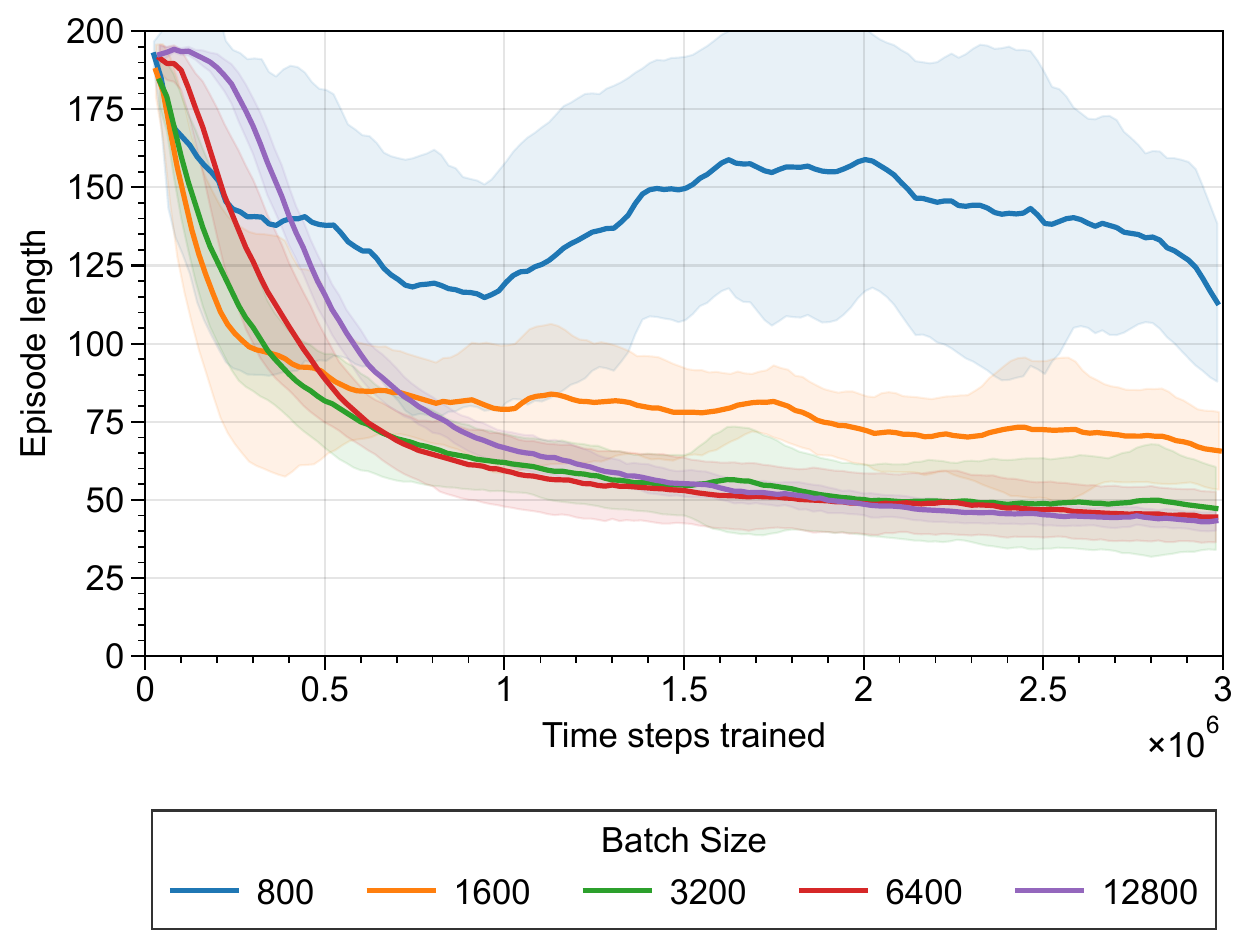}
\caption{Learning curves for agents trained with varying batch sizes.}
\label{fig:batchsize_effects_noaux}
\end{figure}

Training batch size is an important hyperparameter that affects learning dynamics across all branches of machine learning.
It refers to the number of time steps collected by the agent before a gradient update is performed to optimize the agent's network parameters.
Generally, as batch size is increased and more data samples are used to generate a network update, noise in the training samples is averaged out, leading to a more consistent gradient, and hence more consistent improvements during machine learning \cite{mccandlish:2018}.

However, selecting batch size is not as simple as picking the largest size that the training hardware used can handle.
Empirical evidence has shown that in supervised learning with stochastic gradient descent (SGD), overly large batch sizes can negatively impact a model's generalization capabilities \cite{keskar:2016}.
Yet, this phenomenon has not appeared to have the same effect in reinforcement learning tasks \cite{mccandlish:2018, shacklett:2021, stooke:2018}.
In fact, the more common issue with online RL is in using a batch size that is too small, especially in tasks with reward sparsity.
A small training batch may be made up of experiences where no reward is achieved.
In these cases there is no meaningful reward signal that the network can use to update in a useful way.
In RL, any poor training regime becomes exacerbated, as a network that induces a bad policy will in turn collect unhelpful data, leading to a downward spiral of successively worse performance.

Our results are consistent with those previous findings in different RL environments.
Fig \ref{fig:batchsize_effects_noaux} shows learning curves for agents in our simulated Morris Water Maze navigation task, trained with varying batch sizes.
As mentioned in Section \ref{sec:network_training}, training is conducted with 100 parallel copies of the agent collecting in order to collect decorrelated experiences to train with.
A batch size of 1600 for example indicates that each copy collects 16 time steps of experience before a network update is generated.

In Fig \ref{fig:batchsize_effects_noaux}, we observe that agents generally fail to converge to optimal policies when training with small batch sizes (800 and 1600).
However, agents trained with batch sizes 3200 and larger all achieve final performances that are comparable to one another, suggesting that optimal policies can be consistently learned for these sufficiently large batch update sizes.
Note that the 3200 batch size agents converge to optimal policies more quickly than 12800 batch size agents in terms of total time steps of data collected.
This result is consistent with findings from \cite{mccandlish:2018, shacklett:2021, stooke:2018}. 
In particular, McCandlish et al. \cite{mccandlish:2018} suggest that there is a \emph{critical batch size} that describes the number at which larger batch sizes no longer contribute to increased learning efficiency.
This authors suggest that this critical value is dependent on gradient noisiness, which, in the context of RL, increases as environment complexity increases and is especially larger in reward-sparse challenges.

For most of this paper, we consider batch sizes of 1600, often using well-learning 3200 batch size agents as a point of comparison. 
A batch size of 1600 appears to be the regime for our training environment and setup where RL agents can consistently learn to perform the navigation task, but where we may be able to enhance performance with the addition of beneficial auxiliary tasks. 
It should be noted that in simpler environments, such as the one considered here, increasing batch size can be an easy and practical way to improve learning effectiveness. 
However, in complex environments of interest it might be computationally unfeasible to reach the critical batch size. 
For example, McCandlish et al. \cite{mccandlish:2018} suggests that the critical point for environments like Dota \cite{berner:2019} may require batch sizes on the orders of $10^6$ to $10^7$ or greater. 
In such situations we expect auxiliary tasks approach provide an important alternative. 
In the next two sections we introduce a range of auxiliary tasks, conduct a detailed study of their effect on navigation learning, and proceed to consider auxiliary task design.

\subsection{Auxiliary Tasks} \label{sec:aux_tasks}

In this section, we investigate the impact of auxiliary tasks on the performance of our RL agents trained in the North Poster MWM task.
Auxiliary tasks are supplementary objectives the agent must complete alongside the primary RL task.
They can serve to generate meaningful learning signals for the agent, potentially enabling useful updates to neural network weights even in the absence of reward signals.
This is a general description, and many different types of auxiliary tasks have been studied across various RL contexts.

Supervised auxiliary tasks are ones where the agent is tasked with predicting a value about its current state or about the environment.
The true value is given to the agent at network update steps to minimize prediction error with.
Examples include terminal prediction \cite{kartal:2019}, where the agent must predict how many time steps are left in the current episode.
In 3D navigation, supervised auxiliary tasks have been shown to improve learning \cite{lample:2017, mirowski:2016}.
There are also unsupervised auxiliary tasks, which do not use true values to correct prediction errors, such as tasking agents with determining what actions affect the environment \cite{jaderberg:2016}, or with exploration tasks \cite{pathak:2017}.

In the following sections, we also consider what we call reward-based auxiliary tasks (or simply, reward auxiliary tasks).
These tasks directly augment the original reward, and hence do not need modifications to the original RL algorithm to learn.
They can serve to directly address the issue of sparse rewards - a challenge in some RL environments where rewards are rarely given - by consistently offering a dense reward signal to learn from.

Reward auxiliary tasks can be conceived as generalizations to reward shaping \cite{ng:1999}.
However, reward shaping is usually thought of as providing direct guidance towards solving the main goal, while we think of reward auxiliary tasks as providing additional learning signals without requiring direct correlation with the RL task.

In the following subsections, we start by defining a range of supervised and reward-based auxiliary tasks specific to the 2D MWM environment.
Our goal is to determine what types of tasks are beneficial and why.
\\

\subsubsection{Auxiliary task definitions}

We start by defining all auxiliary tasks that are considered in this paper. 
The auxiliary tasks considered can be classified as numerical prediction tasks, categorical prediction tasks, and reward auxiliary tasks.
For brevity, we also call these numerical tasks, categorical tasks, and reward tasks respectively in the text.

\textbf{Numerical prediction tasks.} 
In these tasks, the agent must output a target quantity $\hat{y}_t \in \mathbb{R}$ at every time step.
Outputs are generated from the actor branch of the agent's neural network, as described in Materials and Methods Section \ref{sec:network_training}.
The true value $y_t$ for each of these auxiliary tasks is normalized to be in $[0, 1]$ and given to the agent by the environment during the update step of training.
Tasks are learned by the network minimizing the mean squared error loss function
\begin{equation}
\label{eqn:num_aux_loss}
\mathit{\mathcal{L}_\mathrm{aux,num}} = \frac{1}{N}\sum_{i=0}^N (y_i - \hat{y}_i)^2
\end{equation}
where $N$ represents the total number of time steps in a batch and $i$ indexes each time step.
We use the following numerical prediction tasks:

\textbf{Goal Distance (GD).} 
The agent must output the Euclidean distance between its current position and the center of the goal platform.

\textbf{Angle Distance to Direction (AD).}
The agent must output the shortest angular distance between its current heading and a given direction. 
We use North, East, or both North and East together as target directions.

\textbf{Terminal Prediction (TP).}
This task has been adapted from Kartal et al. \cite{kartal:2019}. 
The agent must output a number indicating how many steps are predicted to remain in the current episode. 
\\

\textbf{Categorical prediction tasks.}
These are also supervised, but require the agent to predict a categorical label, encoded as a one-hot vector $y_t \in \mathbb{R}^d$, where $d$ is the number of classes to predict from.
The task is learned by minimizing the cross-entropy loss function between predictions $\hat{y}_t$ and $y_t$
\begin{equation}
\label{eqn:cat_aux_loss_partial}
\mathcal{L}_\mathrm{aux,cat,t} = -\sum_{i=1}^d y_{t,i}\log(\hat{p}_{t,i})
\end{equation}
\begin{equation}
\label{eqn:cat_aux_loss_full}
\mathcal{L}_\mathrm{aux,cat} = \frac{1}{N} \sum_{t=0}^N \mathcal{L}_\mathrm{aux,cat,t}
\end{equation}
where $\hat{p}_{t,i}$ is the agent's outputted probability that the true label at time $t$ should be $i$.
For our 2D MWM environment, we employ the following tasks:

\textbf{Left Right Turn to Direction (LR).} 
The agent must output a probability vector $\hat{p}_t \in \mathbb{R}^2$ at each time step, indicating whether it is closer to turn left or right to face a given cardinal direction.
This task is analogous to the Angle Distance numerical prediction task.
Similarly, we test this task with North, East, and both North and East directions.

\textbf{Faced Wall (FW).} 
The agent must output a probability vector $\hat{p}_t \in \mathbb{R}^4$ indicating which wall it is currently closest to facing.

\textbf{Quadrant Position (QP).}
The agent must output a probability vector $\hat{p}_t \in \mathbb{R}^4$ indicating which quadrant of the maze it is currently in.
This task is somewhat comparable to the numerical Goal Distance task, where the agent must be aware of positional information.
\\

\textbf{Reward auxiliary tasks.}
As described earlier, reward auxiliary tasks are implemented as augmentations to the RL reward.
We can formally define reward auxiliary tasks with reward functions
\begin{equation}
R^\mathrm{aux}: S \times A \rightarrow \mathbb{R}.
\end{equation}
The agent's new RL task is to maximize combined discounted returns of both the original RL reward and the auxiliary reward
\begin{equation}
G^\mathrm{aux}_t = \sum_{k=0}^\infty \gamma^k (r_{t+k+1} + r^\mathrm{aux}_{t+k+1}).
\end{equation}
It is clear that an agent maximizing $G^\mathrm{aux}_t$ is not guaranteed to learn a policy that maximizes $G_t$. 
In practice, it is important to scale $R^\mathrm{aux}$ values to be much smaller than the original $R$ ones, so that the primary focus remains on completing the original task.
We scale the rewards such that agents are typically at most able to earn a bonus return of $0.1$ across an episode where normal goal navigation is being performed, compared to the reward of $1$ earned for reaching the goal.
We implement two reward auxiliary tasks:

\textbf{Distance Reward (RD).}
The agent is rewarded at each time step proportional to its proximity to the platform.
This reward is linearly scaled such that at the maximum possible distance from the goal there is no bonus reward, and if the agent were to be standing on the center of the goal, it would be rewarded the maximum bonus of $0.0015$.

\textbf{Explore Bonus (RE).}
We divide the water maze into a $5 \times 5$ grid of chunks. 
The agent receives a bonus reward of $0.01$ each time a new chunk is visited on each episode.

In later figures, numerical tasks will be marked by squares ($\blacksquare$), categorical tasks will be marked by crosses ($\times$), and reward auxiliary tasks will be marked by triangles ($\blacktriangle$).

\subsubsection{Auxiliary task learning and performance}

\begin{figure}
\centering
\includegraphics{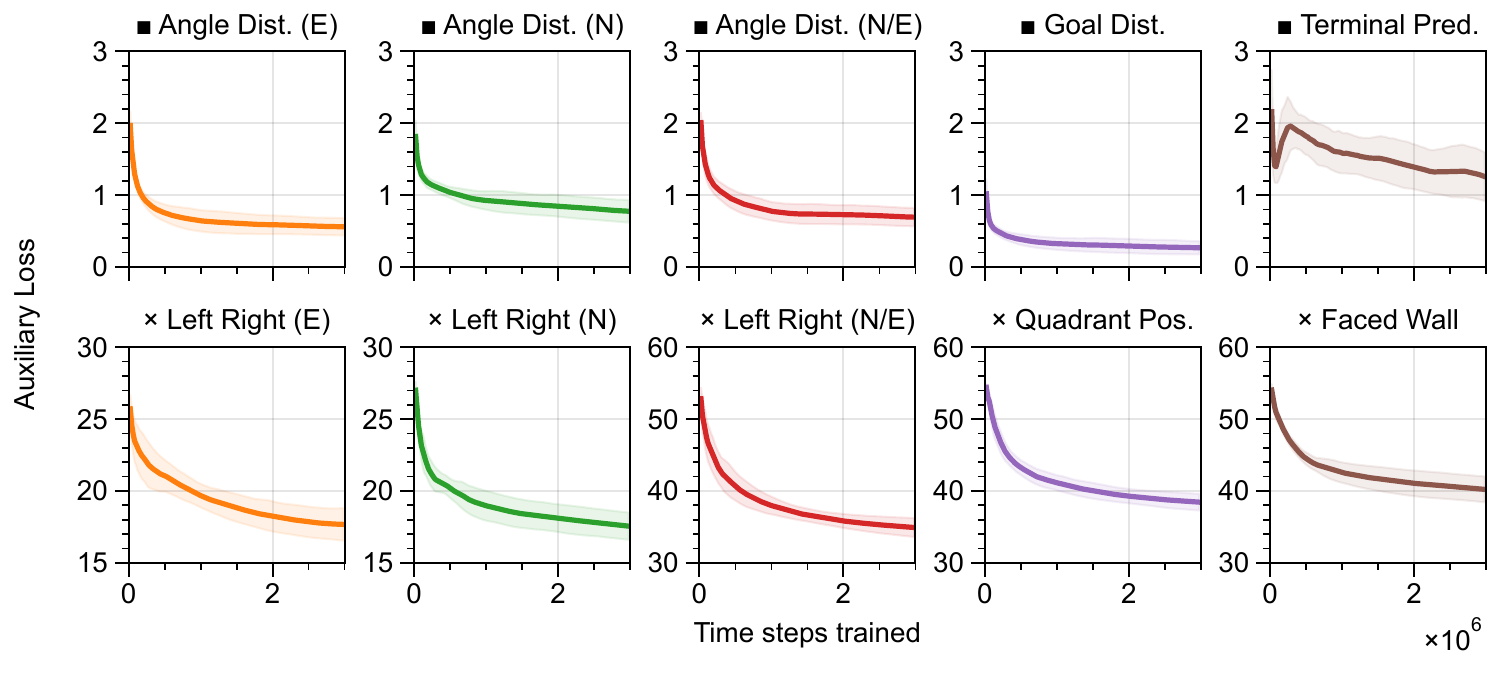}
\caption{Mean auxiliary losses over training period for numerical and categorical auxiliary tasks, averaged across 10 trials.
Only agents trained with a batch size of 1600 are pictures.
Batch size 3200 agents display similar auxiliary loss progression.
\textbf{Top row:} numerical prediction tasks ($\blacksquare$).
\textbf{Bottom row:} categorical prediction tasks ($\times$).}
\label{fig:comb_aux_losses}
\end{figure}

\begin{figure}[t]
\centering
\includegraphics{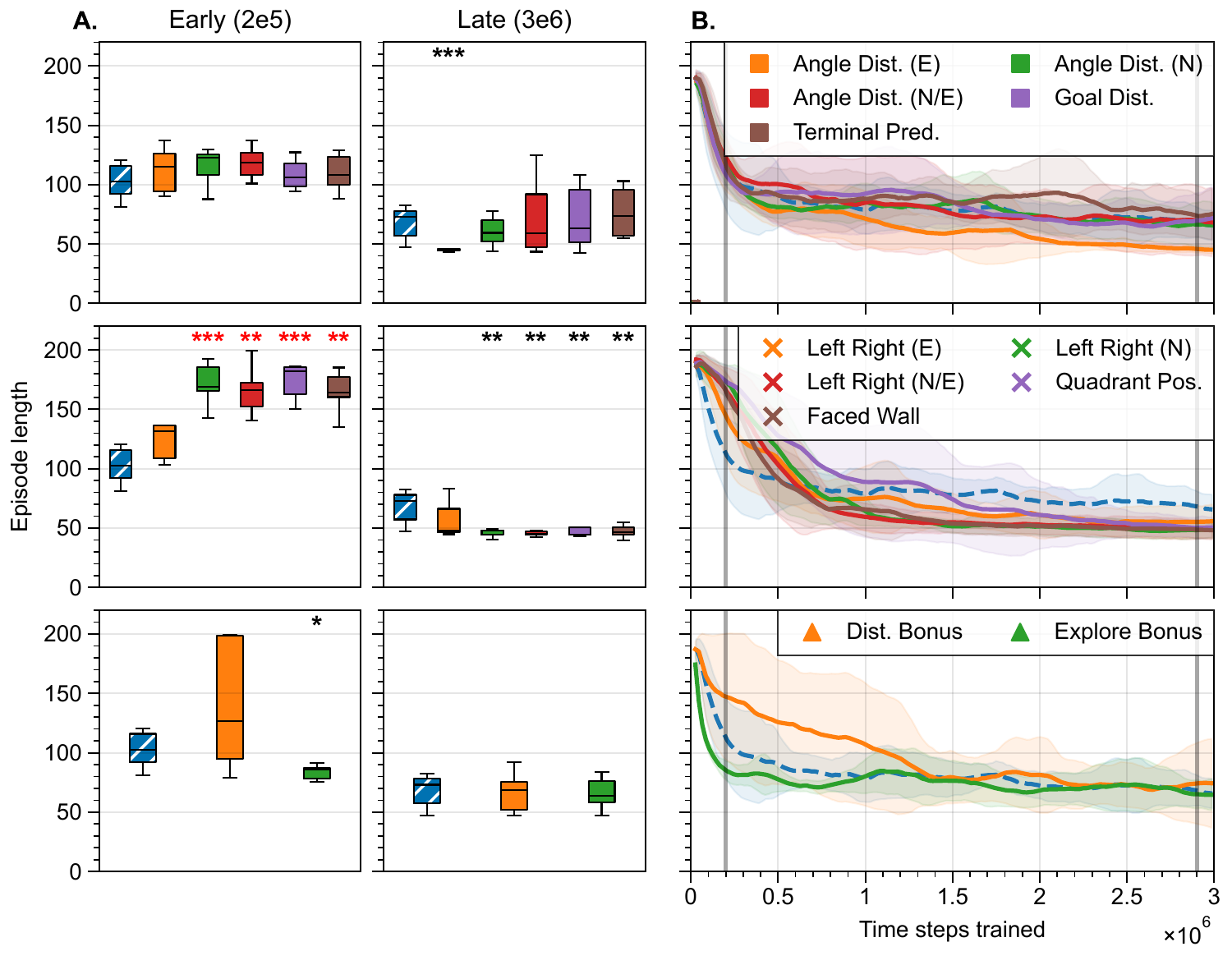}
\caption{
\textbf{A.} Box plots showing median performances of 10 agents (batch size 1600 only) trained with numerical auxiliary tasks at two time points, either early or late in training.
Outliers are not shown.
Control agents (agents trained without auxiliary tasks) are displayed in the left-most blue boxplot with hatches.
Other colors correspond to to different auxiliary tasks as indicated in the legend and in the text.
Stars ($*$) indicate where there is a significant difference between performance of auxiliary task agents and control agents.
Red stars indicate poorer performance, while black stars signify better performance.
The number of stars corresponds to the level of significance ($*$: $p < 0.05$; $**$: $p < 0.01$; $**$\(*\): $p < 0.001$).
\textbf{B.} Training curves of the same agents. 
Lines show the mean performance across 10 agents, and shaded areas showing $\pm 1$ standard deviation.
Control agents are shown as a dashed blue line.
Vertical gray lines indicate the time points used in for box plots in (A).
\textbf{Top row:} Numerical tasks ($\blacksquare$).
\textbf{Middle row:} Categorical tasks ($\times$).
\textbf{Bottom row:} Reward tasks. ($\blacktriangle$)}
\label{fig:comb_aux_performance}
\end{figure}

Next, we examine how agents trained with each of these auxiliary tasks perform, and begin to explore when auxiliary tasks can affect learning rates.
To start, we can confirm that each of these tasks are learnable.
Numerical and categorical tasks induce loss functions that show the error between predictions $\hat{y}_t$ and true values $y_t$.
Fig \ref{fig:comb_aux_losses} visualizes the progression of these losses over training.
All tasks (with the exception of Terminal Prediction) have consistently decreasing losses, confirming that these tasks are being learned.
Numerical tasks notably reach a plateau in losses early in training.
Categorical tasks are learned more quickly at the start than end of training, but agents still improve their ability in categorical tasks across training.

More importantly, auxiliary tasks also influence learning of the main RL task.
These effects are visualized in Fig \ref{fig:comb_aux_performance}.
Fig \ref{fig:comb_aux_performance}B shows the overall training curves for agents trained with each auxiliary task, while Fig \ref{fig:comb_aux_performance}A shows performance specifically at snapshots early and late in training.

Generally, we can see consistent trends throughout each class of auxiliary task.
Most numerical tasks have no significant impacts on RL performance.
Angle Distance (E) is an exception, and this task seems to induce the most consistent improvement over controls all throughout training, compared to all other tasks.
Categorical tasks on the other hand, all improve the final policies that agents converge to, at the cost of slowed early learning.
Although the Left Right (E) does not have the same level of statistical significance as the other tasks, it still qualitatively follows the same trend as seen in the training curves of Fig \ref{fig:comb_aux_performance}B.
Notably, the only auxiliary task which appears to improve early learning rates is the Explore Bonus reward task. 
Intuitively, the Explore Bonus task encourages the agent to move around the play area, increasing its chance of stumbling into the actual goal, which is needed to begin learning usable navigation strategies.

Overall, these results show that categorical auxiliary tasks appear to most broadly improve the policy convergence in the MWM task.
We will now briefly consider the overall differences between numerical and categorical tasks before diving deeper into effects of auxiliary tasks on agent's navigation learning.  
For comparison, consider Angle Distance (numerical) with Left Right (categorical) tasks, which require similar skills for the agent.
The Left Right tasks are easier to succeed on than the Angle Distance ones, as the agent only needs to make a binary prediction in Left Right, rather than output a precise number.
It is possible that if an auxiliary task is too difficult, the agent often predicts $\hat{y}_t$ very incorrectly, generating network updates that are not as beneficial as an easier task.
On the other hand, if tasks are too easy, they may also not provide enough learning signals.
Angle Distance (E) and Left Right (E) were outliers in their respective auxiliary task classes.
We hypothesize that East versions of these tasks may be the easiest to learn, as it is often the direction the agent faces most while navigating towards the goal (possibly related to why the East Poster MWM environment is easier to learn than the North or West Poster variations, as seen in Fig \ref{fig:envtype_learning_curves}).
While most numerical tasks are too difficult, the Angle Distance (E) is just easy enough to perform predictions for.
Conversely, while most categorical tasks are at a sufficient difficulty enough to learn from, the Left Right (E) task is too easy and provides less benefit.
From this perspective, auxiliary tasks that are beneficial for RL may be those in a sweet spot of difficulty, but generally categorical tasks may be closer to this sweet spot and hence less challenging to design well.

In sections that follow, we will focus on categorical tasks and the Angle Distance (E) task as ones that are beneficial in improving late training performance, and less so on the Explore Bonus task which only improves early learning rates.
In particular, we are interested in uncovering the mechanisms of how these auxiliary tasks affect learned policies.

\subsubsection{Auxiliary task behavior classification}

\begin{figure}
\includegraphics{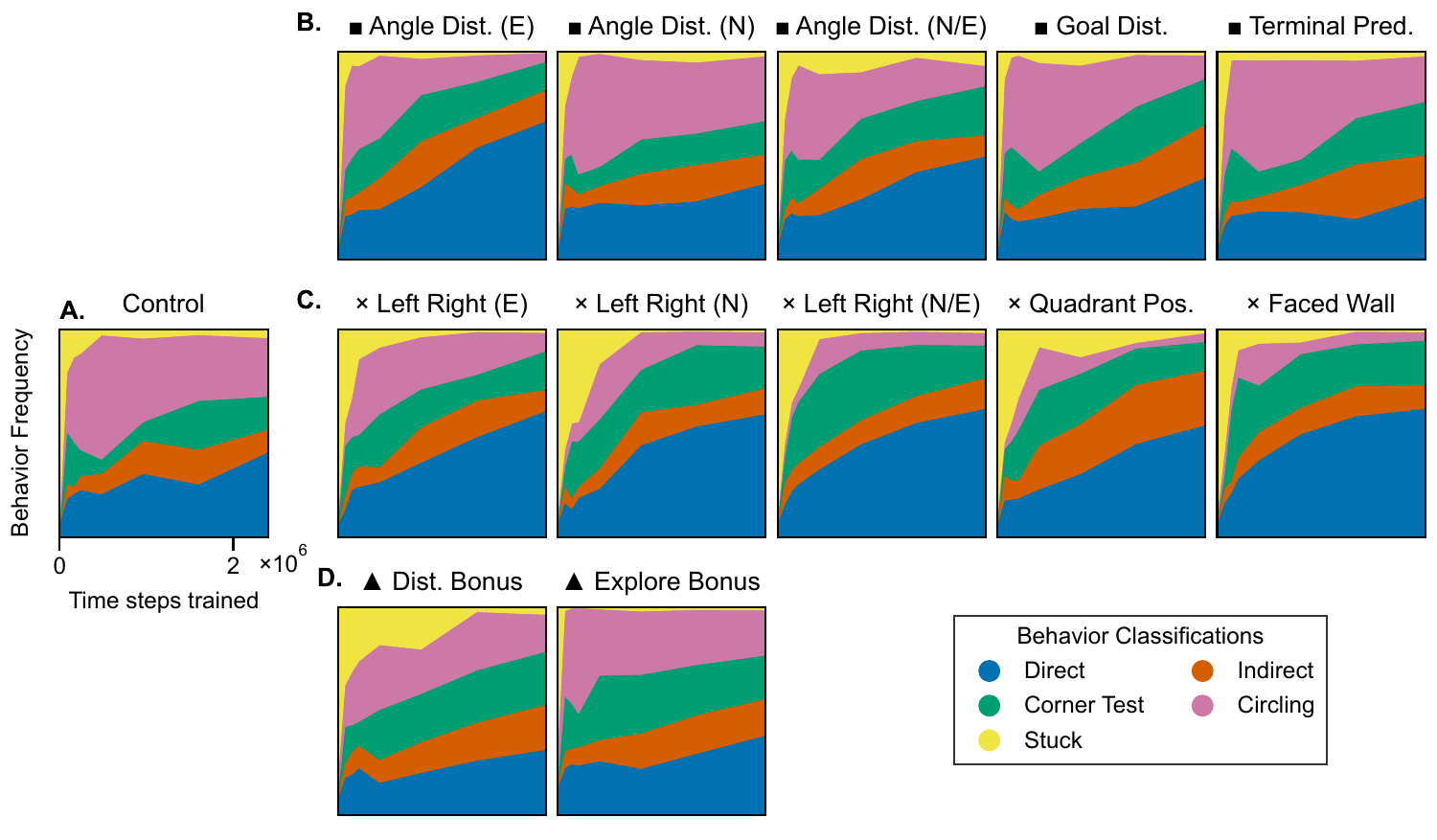}
\caption{Strategy usage across training for agents trained with batch size of 1600 and different auxiliary tasks.
Each subplot shows results of 10 agents collecting 100 episodes each at eight checkpoints through training.
\textbf{A.} Control;
\textbf{B.} Numerical tasks ($\blacksquare$);
\textbf{C.} Categorical tasks ($\times$);
\textbf{D.} Reward tasks ($\blacktriangle$).
Sum of spatial strategies (direct, uncertain direct, corner test) can be seen from the highest green value.
Note that all axes are shared with the one shown for the control.}
\label{fig:aux_task_trajectories}
\end{figure}

In this final section focused on auxiliary task performance, we analyze the navigation strategies used by agents trained with each auxiliary task, as previously done in Section \ref{sec:behavior_analysis}.
These results are shown in Fig \ref{fig:aux_task_trajectories}.
Control agent strategy usage is shown in Fig \ref{fig:aux_task_trajectories}A.
Control agents are able to use spatial trajectories (indicated by the sum of direct, indirect, and corner test areas), but they consistently rely on slower circling methods even towards the end of training.

Qualitatively, assigning auxiliary tasks appears to stabilize the development of strategy usage.
While the amount of direct navigation in control agents fluctuates between checkpoints, almost all auxiliary tasks display steady increases in direct navigation.
Even reward and numerical tasks show consistent increases in spatial strategy usage across training checkpoints, even without inducing statistically significant performance boosts.
The Explore Bonus is especially notable in almost entirely eliminating stuck behaviors even in early training.

\begin{figure}
\centering
\includegraphics{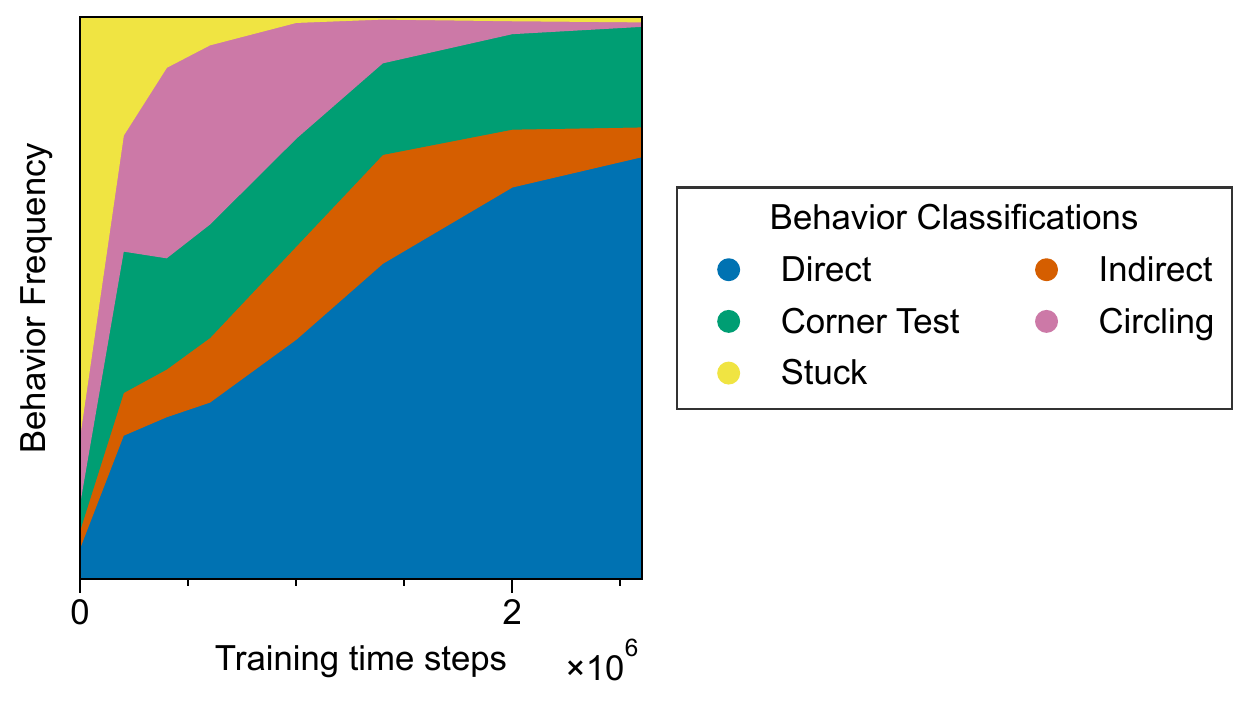}
\caption{Strategy classification of 10 agents trained with a network width of 64 nodes per layer and trained with a batch size of 20000.
We regard these agents as having developed fairly optimal strategies in the North Poster MWM environment.}
\label{fig:64width_trajectories}
\end{figure}

For comparison, we show strategy usage for agents that were trained with network widths of 64 nodes per layer and with a batch size of 20,000 in Fig \ref{fig:64width_trajectories}.
We consider these agents to have learned highly optimal strategies.
The large batch size allows for very consistent network gradients, and the larger neural network size accommodates complex strategy usage.
We note that qualitatively, the progression of strategy usage in these optimal agents resembles those of categorical auxiliary agents, particularly the Faced Wall and Left Right (N/E) ones.

Perhaps most notably, categorical task agents (Fig \ref{fig:aux_task_trajectories}C) which have improved performance of final policies appear to almost entirely eliminate their usage of circling strategies.
In fact, many of these agents appear to barely develop any circling strategies, even in early training (with Left Right (E) agents being an exception), and even in comparison to batch size 20,000 agents.
We can infer that assigning categorical tasks enhances performance by preventing reliance on simple circling behaviors and encouraging the development of spatial navigation methods.
However, this comes at the cost of more agents being stuck in early checkpoints rather than relying on ``easy'' circling strategies.
As a consequence, they have decreased early performance.
We explain the influence of categorical auxiliary tasks on strategy usage further after analyzing how auxiliary tasks affect representation development in Section \ref{sec:rep_strategy_correlation}.

\subsection{Gradient Cosine Similarities} \label{sec:grad_cos_sim}

Each time a neural network is updated via stochastic gradient descent, we can think of the neural network weight updates as a vector pointing in the direction of greatest descent of a loss function, or the direction of greatest ascent for policy gradient RL methods.
We call this vector the gradient, and $\nabla_\theta \mathcal{L}_\mathrm{RL}$ is the gradient induced by the RL task.
Du et al. \cite{du:2018} suggest considering the cosine similarity between $\nabla_\theta \mathcal{L}_\mathrm{RL}$ and any gradients induced by auxiliary tasks $\nabla_\theta \mathcal{L}_\mathrm{aux}$.
They proposed that when these gradients have positive cosine similarity, one might expect that the auxiliary task is beneficial to learning the RL task.
Lin et al. \cite{lin:2019} used this idea to create an algorithm that adaptively weights auxiliary gradients in the update step based on cosine similarity.
It has been suggested \cite{du:2018, lin:2019} that evaluating auxiliary tasks based on their induced gradient vectors can help mitigate potential learning penalties incurred by auxiliary tasks.  

It is important to note that naively requiring auxiliary gradients to have positive cosine similarity to RL gradients during training may not always be optimal.
Consider for example, a newly initiated RL agent with an effectively random policy.
This agent has low probability of reaching the goal and generating a useful reward signal to learn from.
Under these circumstances, the agent may still be able to develop useful representations of the environment through auxiliary tasks.
In such cases, it may be desirable to apply auxiliary gradients even if they have low similarity with the RL gradients.
Next, we will evaluate the gradients associated with various auxiliary tasks, their alignment with the RL gradients, and the relationship with the task's effectiveness. 

\subsubsection{Supervised auxiliary task gradients}

\begin{figure}
\centering
\includegraphics{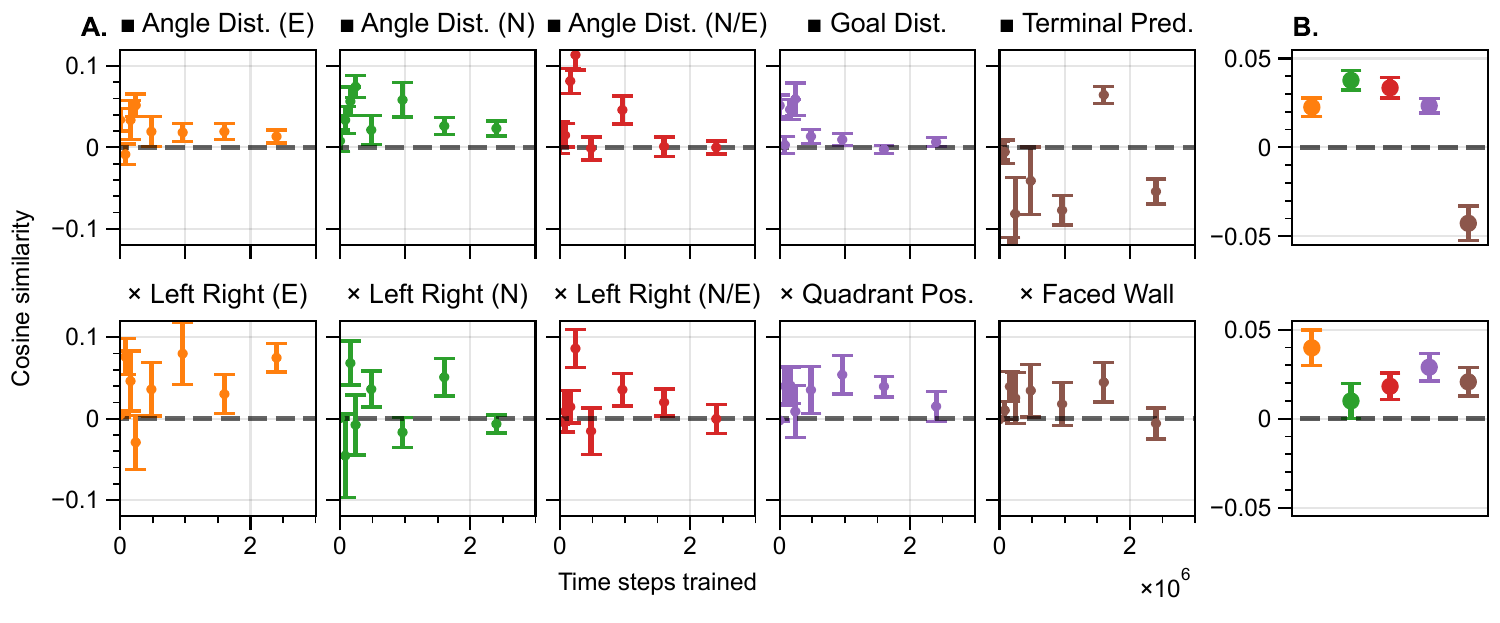}
\caption{
\textbf{A.} 95\% confidence intervals for cosine similarities of each supervised auxiliary task, across 10 agents and across eight checkpoints in training.
Dotted line shows where $0$ is on the y-axis.
\textbf{B.} 95\% confidence intervals for all checkpoints.
Colors correspond to each auxiliary task shown in (A).
\textbf{Top:} Numerical tasks ($\blacksquare$).
\textbf{Bottom:} Categorical tasks ($\times$).
}
\label{fig:cos_sim_checkpoints}
\end{figure}

We describe the process for computing supervised auxiliary task gradients and cosine similarities in the Materials and Methods Section \ref{sec:methods_aux_gradients}.
Fig \ref{fig:cos_sim_checkpoints}B shows  these confidence intervals for cosine similarity averaged across the eight checkpoints.
All auxiliary tasks on average have positive cosine similarity with the RL task, except for the Terminal Prediction numerical task, which has negative mean cosine similarity.
However, the positive values are very small, and there seems to be little significant difference between cosine similarity measures across different auxiliary tasks, and no clear correlaation between mean cosine similarity and actual effects on training performance.
For the most part, we observe that using cosine similarity as a measure for how much a task may benefit learning the RL goal in inconclusive.
It is notable that in our MWM task, categorical auxiliary tasks provide significant benefit to final learned policies despite having near $0$ cosine similarity (Fig \ref{fig:comb_aux_performance}B).

\subsubsection{Reward auxiliary task gradients}

Calculating reward auxiliary task gradient cosine similarities is different than the process for supervised tasks, and is also describe in Section \ref{sec:methods_aux_gradients}.
It is important to note that in the actor-critic algorithm, the agent learns by determining which actions lead to better than expected returns.
In the pure bonus gradient, we are artificially decreasing rewards compared to that expectation.
As an agent develops a better performing policy, it expects more rewards, causing the pure bonus gradient and RL gradient to diverge more.

\begin{figure}
\centering
\includegraphics{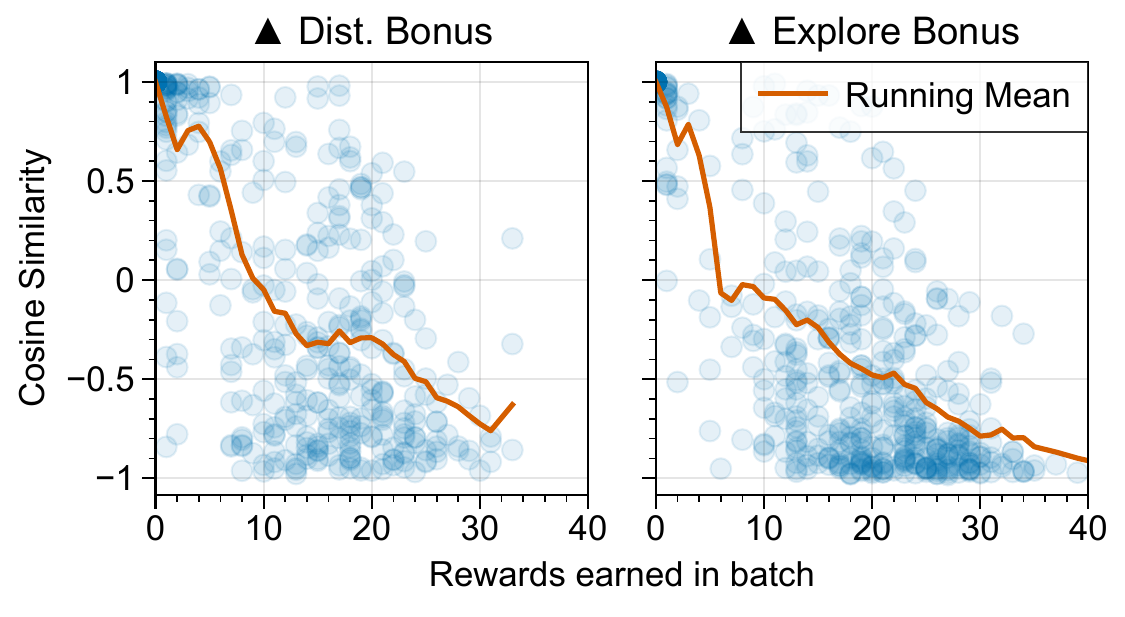}
\caption{Cosine similarities between RL gradients and pure reward gradients, organized by the number of goal rewards earned in the batch.
Blue dots indicate the cosine similarities for individual batches.
The orange line shows a running exponential mean of the cosine similarity measure.
Only agents trained with 1600 batch size are shown.}
\label{fig:reward_cos_sims}
\end{figure}

This divergence is demonstrated in Fig \ref{fig:reward_cos_sims}.
Instead of organizing cosine similarity by time steps trained as we did in Fig \ref{fig:cos_sim_checkpoints}A, we organize it by the number of goal rewards that were earned in each particular batch.
This shows the clear divergence of gradients mentioned.
We see that for most ranges of performance there is negative cosine similarity between $\nabla_\theta \mathcal{L}_\mathrm{RL}$ and $\nabla_\theta \mathcal{L}_\mathrm{bonus}$.
Similar to supervised tasks, there seems to be no significant difference of cosine similarity measures between the two reward auxiliary tasks, despite differences in effects on learning.
We conclude that cosine similarity in general is ineffective for distinguishing usefulness of auxiliary tasks in our RL setup.
Conversely, this demonstrates that auxiliary tasks can be useful and help agents learn about the environment even when their gradients do not closely align with the gradients induced by the main RL task.

\subsection{Representation Analysis}
\label{sec:representation_analysis}

In the final results section, we investigate the ``representations'' developed by RL agents while learning the 2D MWM task. 
The representations are defined formally in Materials and Methods Section \ref{sec:reinforcement_learning} as functions that map observations $o_t$ and internal states $h_t$ to multi-dimensional vectors. 
Practically, the representations can be thought of as snapshots of activity in the agent’s networks at the time $t$ when an agent is in the state $h_t$ and is observing $o_t$.  
Our inspiration in considering representations comes primarily from recordings of neural activity in rodents’ brains during navigation. 
For example, place cells and grid cells have long been described and studied in the hippocampus and entorhinal cortex, respectively. 
These neurons exhibit firing rates which are sensitive to the rodent’s position in space \cite{moser:2015}. 
Place cells fire preferentially at single specific locations, while grid cells fire periodically in space. 
There are also head-direction cells found in multiple cortices of rodents \cite{taube:1990}. 
These neurons have firing rates which are sensitive to the current direction the rodent is facing rather than the animal’s position.

We imagine that representations like those naturally observed in rodents’ brains could be especially relevant in the context of the simulated North Poster MWM environment. 
At each time step, RL agents receive observations $o_t$ representing visual input, which cannot be used in general to uniquely determine the current state $s_t$. 
However, the environment state is uniquely described by the agent’s position and the direction it faces. 
It is conceivable that if an agent had access to both location and direction information coded explicitly in its networks, it could generate an effective navigation strategy to reach the goal. 
Thus, we are going to explore whether representations resembling those corresponding to place cells or direction cells appear in the navigating agents, and if so – under what circumstances, and what strategies they correlate with.

\subsubsection{Uncovering representations}

As mentioned in Section \ref{sec:reinforcement_learning}, we can conceptualize representations as functions that map observations $o_t$ and internal states $h_t$ to d-dimensional vectors.
For our agent's neural network, we treat the activation of each node in the network as being a component of this feature vector.
We also refer to an individual node as having a spatial or angular representation if its activation pattern is spatially or angularly sensitive during normal agent behavior.

Prior work has formally considered representations as being d-dimensional features $\phi(s_t) \in \mathbb{R}^d$ that are used as a weighted linear combination to approximate $V(s_t) = \phi(s_t)^Tw$ where $w \in \mathbb{R}^d$ is a weight vector \cite{bellemare:2019, lyle:2021}.
In the context of using neural networks to approximate $V$, it is natural to think of these features as the activations of the final layer of the network (which are used in weighted sums to produce $V$ or $\pi$).
However, qualitatively we find the most visually compelling representations in the first fully-connected actor layer after the recurrent layer (see Fig \ref{fig:mwm_poster_env_nn}B) and this is the layer we will focus on for the rest of this section.

\subsubsection{Spatial representations}

\begin{figure}
\centering
\includegraphics{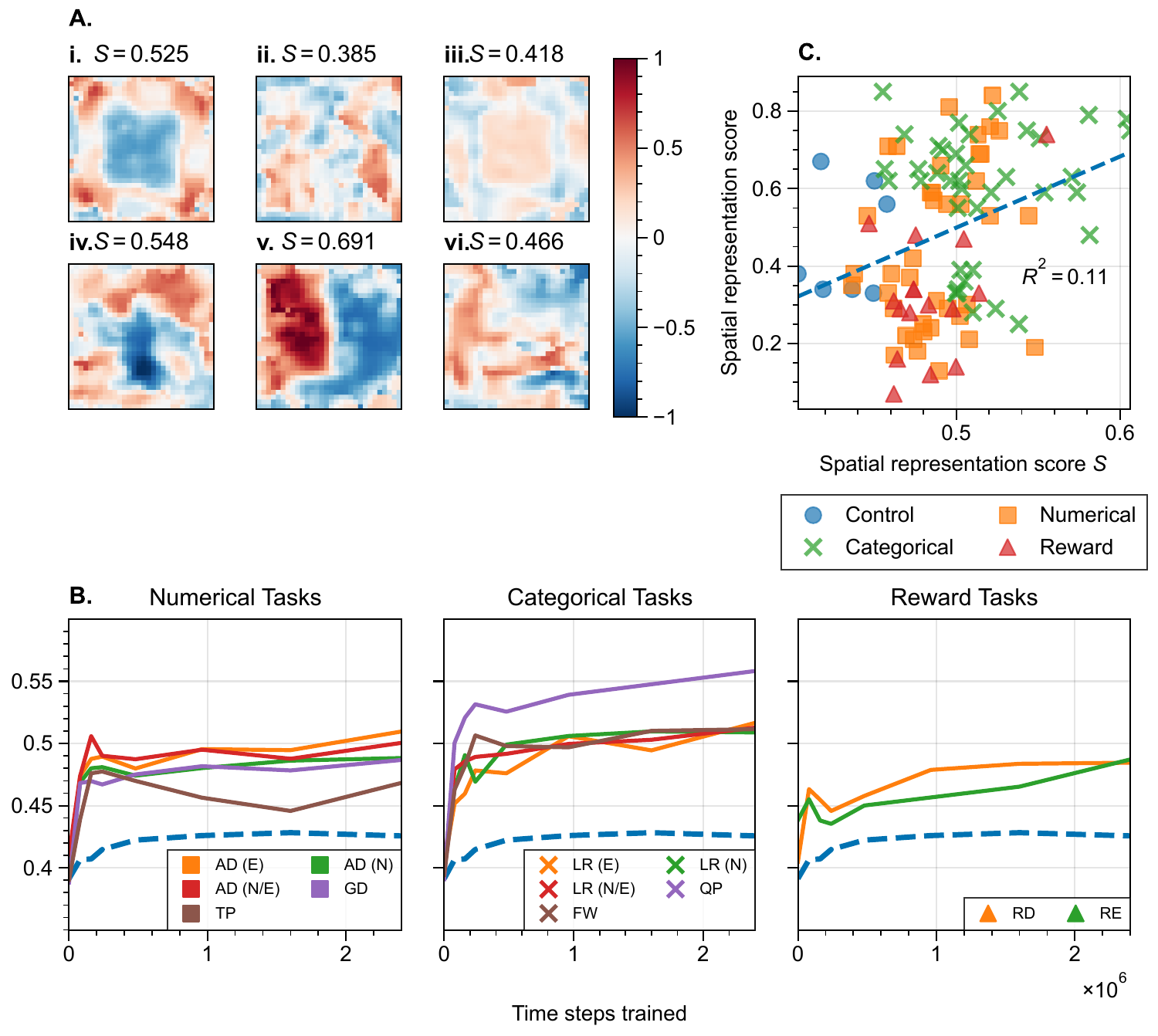}
\caption{
\textbf{A.} Example spatial representation heatmaps from agents trained on different tasks and at different points in training.
Each heatmap comes from a single node of a single agent. 
The color shows the value of $\tilde{a}_i$, where red indicates above average activation and blue indicates below average activation (see Materials and Methods).
The title above each subplot shows the spatial representation score of the heatmap.
\textbf{i.}  Control, 3200 batch size (subplots ii-vi use 1600 batch size), $1e5$ trained time steps.
\textbf{ii.} Control, $2e5$ trained time steps.
\textbf{iii.} Angle Distance (E) task, $1e5$ trained time steps.
\textbf{iv.} Goal Distance task, $2.5e6$ trained time steps.
\textbf{v.} Quadrant Position task, $2.5e6$ trained time steps
\textbf{vi.} Left Right (E) task, $2.5e6$ time steps.
\textbf{B.} Mean spatial representation scores for 1600 batch size agents across eight checkpoints in training.
Scores for control agents are shown with a dashed blue line.
Abbreviations are as follows: 
AD: Angle Distance. GD: Goal Distance. TP: Terminal Prediction.
LR: Left Right. QP: Quadrant Position. FW: Faced Wall.
RD: Distance Bonus. RE: Explore Bonus.
\textbf{C.} Comparison between spatial representation score and `direct' strategy usage of agents late in training ($2.5e5$ time steps). 
Each point represents an individual agent.
The dashed line shows the linear line of best fit with corresponding $R^2$ value.
}
\label{fig:spatial_rep_combined}
\end{figure}

The process for calculating spatial heatmaps $\tilde{a}_i$ is given in Materials and Methods, Section \ref{sec:methods_representation_maps}.
Fig \ref{fig:spatial_rep_combined}A provides a few examples of what these $\tilde{a}_i$ node activation heatmaps look like plotted in 2D space.
Note that these heatmaps are created individually for each node in the network, and also depend on trajectories taken by the agent. 
Each heatmap shows regions in space where the node was more (red) or less (blue) active than average. 

These heatmaps have clear dependence on the actual trajectories that the agents follow.
For example, Fig \ref{fig:spatial_rep_combined}Ai-iii all feature agents primarily performing circling strategies, which is evident in the shape of heatmaps in i and iii.
Some heatmaps may exhibit spatial periodicity (Fig \ref{fig:spatial_rep_combined}Ai), while others display a more distinct local spatial preference (Fig \ref{fig:spatial_rep_combined}Aiv and v).
Nodes from control agents (Fig \ref{fig:spatial_rep_combined}Aii) in particular show some of the least spatially coherent representations.

To quantify a notion of quality in these representations, we develop a spatial representation score calculated as follows.
First, consider the data set of all positive $\tilde{a}$ activations on the grid.
We compute
\begin{equation}
S_+ = \sum_{ij, \; \tilde{a}_i > 0, \; \tilde{a}_j > 0} \exp(-d(x_i, x_j)/\sigma)(\tilde{a}_i \tilde{a}_j)
\end{equation}
for each $i,j$ pair in the data set. Here,
$d(x_i, x_j)$ is the Euclidean distance between points, which we weigh exponentially with $\sigma=50$.
This sum assigns greater weight to closer pairs of points and pairs with higher activations.
We perform the same procedure for negative $\tilde{a}$ grid activations to get $S_-$, and the final spatial representation score is given by
\begin{equation}
S = \frac{S_+ + S_-}{\sum_i \tilde{a}_i^2}.
\end{equation}
This score is designed to assign higher scores when positive and negative areas of activation are well-separated, and when large magnitude activations of the same polarity are close together.
The normalization in the denominator reduces the likelihood of a heatmap scoring highly simply due to having polarized activations, rather than having interesting spatial structures.
Alternative methods for defining a spatial representation score could emphasize different attributes or types of heatmap.
Fig \ref{fig:spatial_rep_combined}A shows spatial representation scores $S$ above each heatmap.

Fig \ref{fig:spatial_rep_combined}B shows the mean spatial representation score averaged over all nodes for 10 agents across 8 checkpoints in training. 
Different colors correspond to control agents or those performing different auxiliary tasks.
Spatial scores qualitatively appear to correlate with performance metrics seen in Section \ref{sec:aux_tasks}.
Categorical tasks consistently improve the development of spatial representations compared to control agents.
Numerical tasks also slightly encourage spatial representation development, and the Angle Distance (E) task, which had significant final performance improvement over the control, has the highest final spatial score among numerical tasks.
Quadrant Position task agents develop the strongest spatial representations among categorical tasks, which may be expected as this task requires agents to have the most positional awareness.

\subsubsection{Direction representations}

\begin{figure}
\centering
\includegraphics{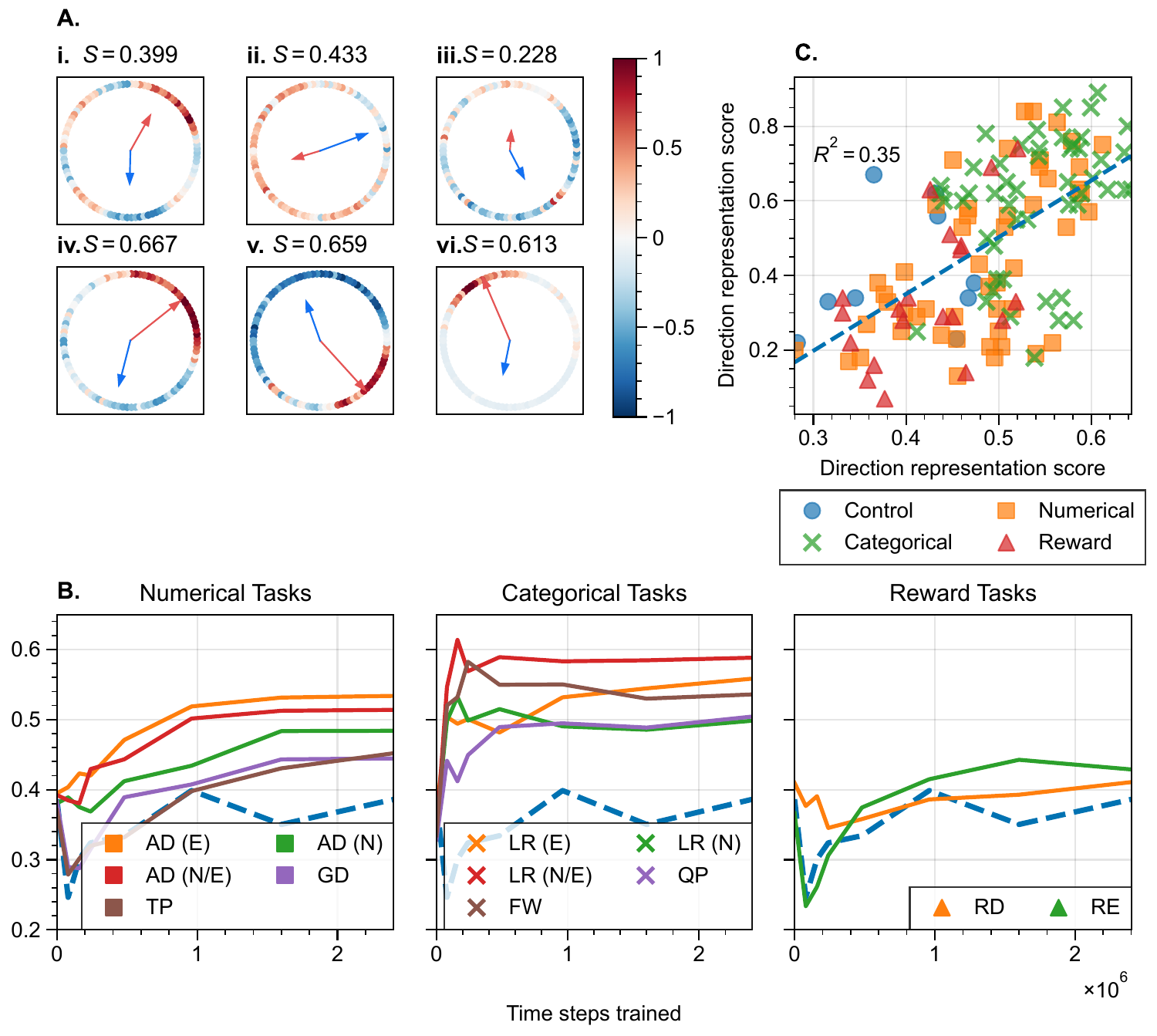}
\caption{\textbf{A.} Example direction maps from agents trained on different tasks and at different points in training. 
Each map comes from a single node of a single agent, and all examples are from agents trained with 1600 batch size.
The color of dots shows the value of $\tilde{a}_i$, with red indicating above average activation and blue indicating below average activation.
Arrows inside each circle show the mean resultant directions and length of positive (red) and negative (blue) activations, and the title above each subplot shows the mean resultant length, which is also the direction representation score.
\textbf{i.} Control, $5e5$ trained time steps. 
\textbf{ii.} Angle Distance (E) task, $1e5$ trained time steps.
\textbf{iii.} Goal Distance task, $2.5e6$ trained time steps.
\textbf{iv.} Quadrant Position task, $2.5e6$ trained time steps. 
\textbf{v.} Left Right (E) task, $2.5e6$ trained time steps.
\textbf{vi.} Faced Wall task, $2.5e6$ trained time steps.
\textbf{B.}
Mean resultant lengths for 1600 batch size agents across eight checkpoints in training for each auxiliary task.
Scores for control agents are shown with a dashed blue line.
Abbreviations are as follows: 
AD: Angle Distance. GD: Goal Distance. TP: Terminal Prediction.
LR: Left Right. QP: Quadrant Position. FW: Faced Wall.
RD: Distance Bonus. RE: Explore Bonus.
\textbf{C.}
Comparison between direction representation score and `direct' strategy usage of agents late in training ($2.5e5$ time steps). 
Each point represents an individual agent.
The dashed line shows the linear line of best fit with corresponding $R^2$ value.
}
\label{fig:direction_rep_combined}
\end{figure}

We employ a similar method to generate maps for direction representations as we did for spatial representations, again as described in Section \ref{sec:methods_representation_maps}.
Visual examples of direction maps and their corresponding direction representation scores are shown in Fig \ref{fig:direction_rep_combined}A.

To quantify the quality of direction representations, we use a fairly natural measure called the mean resultant length.
From direction maps, we first collect all positive $\tilde{a}_i$ grid activations.
Each grid point is converted into a vector with direction given by its angle and length given by $\tilde{a}_i$.
The mean direction and length of these vectors is known as the mean resultant direction and length, respectively.
The same is computed for negative $\tilde{a}_i$ activations.
In Fig \ref{fig:direction_rep_combined}A, the resultant lengths are represented independently by red and blue arrows, and the mean length of these lengths is given in the title of each subsubplot as $S$, the direction representation score. 

Network nodes that consistently activate or deactivate when the agent faces specific directions will have long resultant vectors (Fig \ref{fig:direction_rep_combined}iii-vi), which we think of as strong or good representations.
Those with less specific or consistent direction-based responses have shorter vectors, (Fig \ref{fig:direction_rep_combined}i-iii).
It is worth noting that this measure does not account for potential periodic direction representations.
For example, Fig \ref{fig:direction_rep_combined}i shows a node that appear to activate periodically as a function of faced direction but has short resultant lengths that are not indicative of its consistency.

Fig \ref{fig:direction_rep_combined}B presents the mean of both positive and resultant lengths for agents trained on each auxiliary task across training. 
We observe that for 1600 batch agents, all auxiliary tasks except for reward-based ones improve the development of direction representations.
Tasks with a focus on heading, in particular, tend to be the best (Angle Distance, Left Right, Faced Wall).
Among numerical tasks, the Angle Distance (E) task promotes the best development of both spatial and direction representations.
Notably, this was also the numerical task that led to significant performance improvement over control agents.

\subsubsection{Correlations between representations and strategy usage}
\label{sec:rep_strategy_correlation}

We visualize the relation between  spatial (Fig \ref{fig:spatial_rep_combined}C) and direction (Fig \ref{fig:direction_rep_combined}C) representation scores of individual agents with the frequency that they employ the `direct' navigation strategy late in training.
Naturally, there is a strong positive correlation between frequency of direct navigation and performance on the MWM task, so this plot also correlates representation scores to performance.
While both scores show some positive correlation with direct navigation usage, the direction representation score appears to have a stronger correlation than the spatial score.
This may be partially dependent on the definitions of each score, but it might also indicate that for an agent in the MWM environment, knowing its faced direction is more important than knowing its spatial location.
This could be due to the fact that the platform is always in the South-East corner of the maze, making the ability to consistently head South-East an important part of a successful navigation strategy in this context.
Given a different task, we might find better performance correlation between spatial representation scores than direction ones.

Interestingly, both spatial and direction representations seem to develop early in training without much change after around $10^6$ time steps of training (Fig \ref{fig:spatial_rep_combined}B and Fig \ref{fig:direction_rep_combined}B).
This suggests that representations develop fairly early in training, and further policy improvements occur as the agent network optimizes to incorporate understanding of the environment into decision making.
The period of early representation development also corresponds with the period where most auxiliary task learning occurs, as seen by the steeper earlier auxiliary loss decreases in Fig \ref{fig:comb_aux_losses}.
Referring back to the strategy usages of agents trained under different tasks in Fig \ref{fig:aux_task_trajectories}, we hypothesize that improved representations made available early in development by learning auxiliary tasks make it easier for agents to develop complex navigation strategies.
At the same time, network update gradients from the RL and auxiliary task may be in competition, preventing agents from optimizing their policies as quickly during early training.
This may contribute to the decreased early usage of easier circling strategies seen in categorical auxiliary task agents.

\subsubsection{Combined faced wall and quadrant position task}

\begin{figure}[t]
\centering
\includegraphics{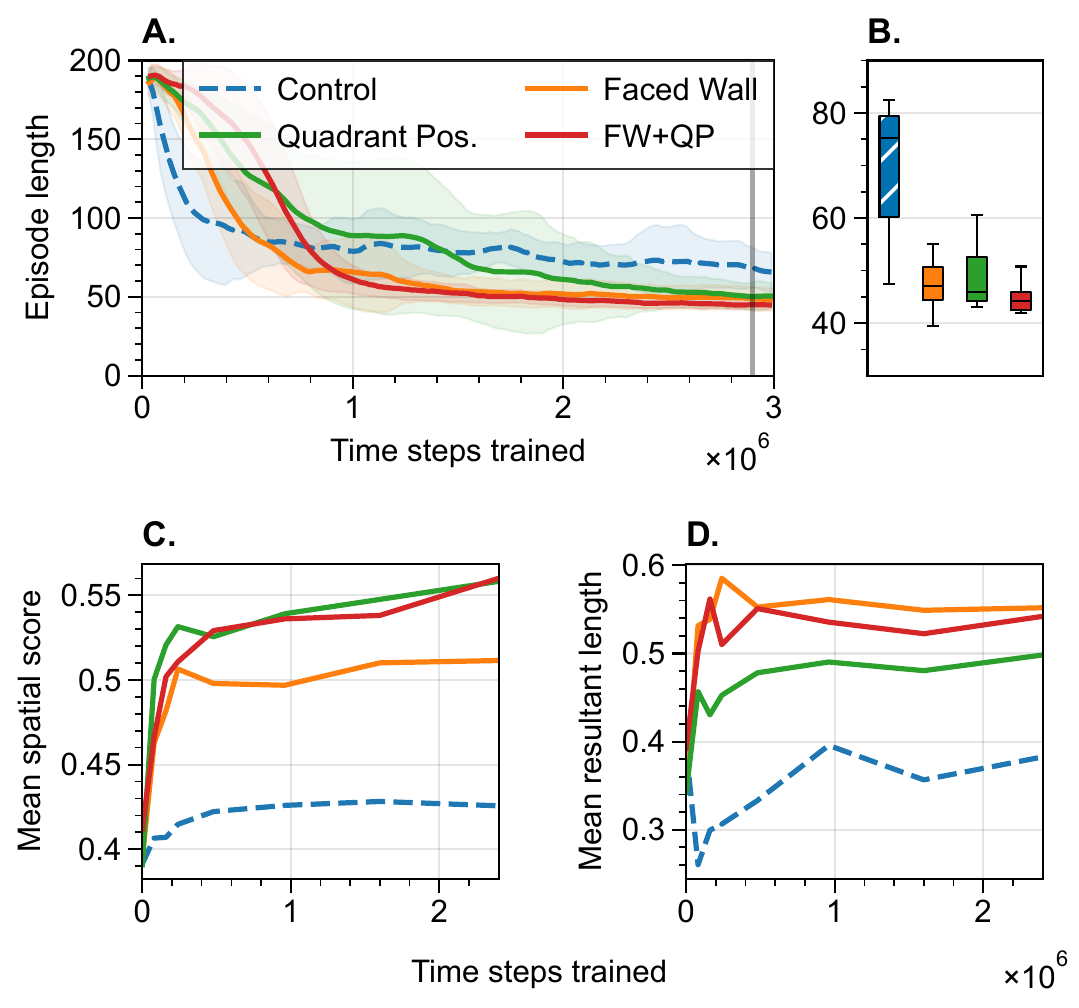}
\caption{\textbf{A.} Training curves for 10 agents trained with 1600 batch size. 
Solid lines show mean performance and shaded areas show $\pm 1$ standard deviation.
\textbf{B.} Box plot showing performance late in training.
\textbf{C.} Spatial and \textbf{D.} direction representation scores for combined FW+QP task compared to independent auxiliary tasks.}
\label{fig:fwqp_comb}
\end{figure}

We finally explore whether we can specifically aim to improve representation scores with combinations of auxiliary tasks.
To achieve this, we train agents with the Faced Wall and Quadrant Position auxiliary tasks simultaneously (we call this the FW+QP task), as these are two of the best-performing tasks in terms of direction and spatial representation development, respectively.

The resulting performance of the combined task is shown in Fig \ref{fig:fwqp_comb}A-B.
As one might expect from previous results, the combined task appears to further slow down early training compared to assigning either individual auxiliary task.
Late in training, these agents demonstrate better performance in both mean and median than the individual tasks.
The difference is not statistically significant but still notable, with the FW+QP task having better performance than either the Faced Wall ($p = 0.21$) or Quadrant Position ($p=0.12$) alone.
The representations scores for the FW+QP agents are shown in Fig \ref{fig:fwqp_comb}C-D.
While these agents do not develop better representations than either individual task agents do, they achieve both spatial and direction representations that are comparable to the best of either task.

\subsection{Comparisons to biological agents}

\begin{figure}
\centering
\includegraphics{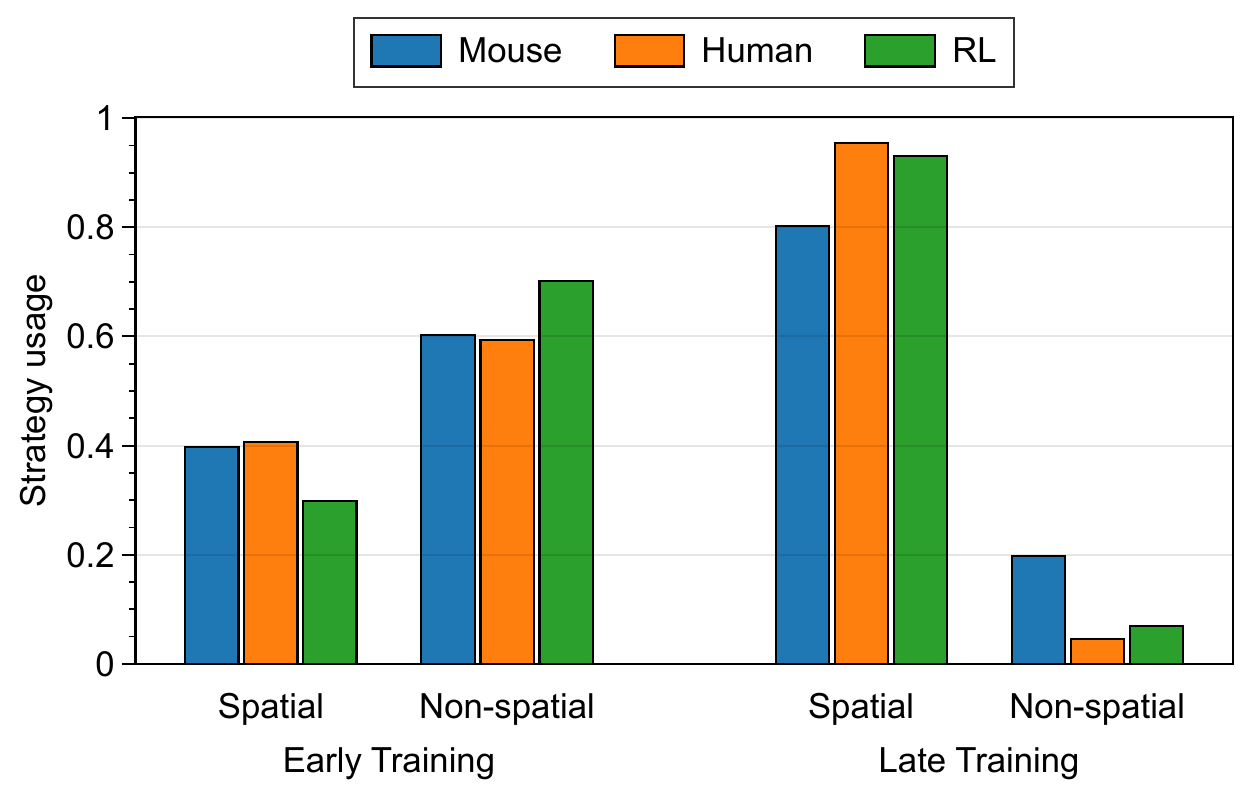}
\caption{Comparisons of behavior usage early and late in training between our RL agents, mice in real MWMs and humans in virtual MWMs. Mouse data comes from Fig 1d in Overall et al. \cite{overall:2020} and human data comes from Fig 5 (4-days protocol) in Schoenfeld et al. \cite{schoenfeld:2017}}
\label{fig:behavior_comparisons}
\end{figure}

Despite the differences between an artificial MWM task with RL agents and real world experiments with mice and humans, we can still observe similarities in each of their learning dynamics.
Fig \ref{fig:behavior_comparisons} shows navigation strategies used across species, compared with RL agents.
Data for mice \cite{overall:2020} and humans \cite{schoenfeld:2017} are taken from early (the first trial) and late (the second-to-last trial) stages of training.
Because it is difficult to compare exact navigation strategies between biological agents and artificial ones, we only consider whether strategies were spatial or non-spatial.
In mice data thigmotaxis, random search, and scanning are non-spatial strategies, while chaining, directed search, focal or corrected search, and direct paths are spatial ones \cite{garthe:2009}.
In humans floating, thigmotaxis, and scanning are non-spatial strategies, while focal search and direct navigation are spatial ones \cite{schoenfeld:2017}.
For RL data, we choose agents trained in the North Poster MWM environment with the Faced Wall auxiliary task.
Early training is defined as 80,000 time steps of training, and late is defined as 1,600,000 time steps.
Direct, indirect, and corner test are spatial strategies, while stuck and circling are non-spatial.

All species and agents show similar patterns of developing more effective spatial navigation skills as they spend more time in their respective water mazes.
Similar trends of decreasing escape latency can also be seen with increased experience.

At the same time, there are some key differences between our RL agents and biological ones.
Firstly, regardless of the auxiliary task, our RL agents always show some amount of `circling' behavior during the intermediate stages of training, and sometimes at the end of training as well.
The same behavior can be seen in mice and humans, and is typically called `chaining' \cite{garthe:2009}, characterized by swimming in circles around the pool at a fixed distance from the walls.
While this appears to be an effective strategy to develop early in training, it is not very prevalent in mice and humans \cite{overall:2020, sandstrom:1998, schoenfeld:2017}.
This highlights an important difference in how agents learn.
Artificial agents are optimized with a single objective of reaching the goal, and any effective strategy to do so can be exploited by its policy.
On the other hand, animals and humans may have other internal motivations that shape behavior and learning.
For example, mice have a preference to stay near walls for safety \cite{garthe:2009}, and may treat the platform as an escape from the dangers of swimming \cite{whishaw:2000}, leading to a preference for direct trajectories.
The RL algorithm naturally optimizes for faster trajectories as well, but only because doing so increases expected long-term rewards earned, which one might consider a weaker learning pressure than an animal might have.

Another important difference is in the shape of the mazes. 
While MWM experiments typically use circular pools, we opted for a square maze, with the intent to modify it for future studies that are easier to design in a square maze.
Even so, most behaviors that we observe could be performed in a circular maze too.
For example, the shape of `circling' trajectories employed by agents approximates a circle, suggesting that if the same agents were in a circular maze, they might still develop some of the same strategies.

The `corner test' strategy may be an exception to this, and likely develops due to both the shape of the maze and limitations on moving and turning simultaneously, as discussed in Section \ref{sec:behavior_analysis}.
This can be argued to be similar to the `landmark' strategy recorded in humans, where a navigator first walks to a known landmark before correcting to the goal \cite{astur:2004, schoenfeld:2010}.
The corners of the square water maze act as landmarks for navigation, breaking symmetries that would be seen in circular environments.
The `corner test' that we see in RL agents could be considered a method of first navigating to a salient landmark as well as performing a guess-and-check.

Finally, we consider whether RL agents in this setting are more similar to rodents or humans.
Whishaw et al. \cite{whishaw:2000} demonstrated that rats performing the same Morris navigation task in a dry setting act differently than when swimming.
They attributed this to swimming in a pool invoking fear, leading to escape navigation, whereas in the dry land case, rats behaved more like they were foraging.
We argue that both humans and our RL agents act more similarly to dry land rodents, as they do not have natural aversion or punishment for time spent in the maze, instead being driven primarily by potential reward in reaching the goal.
If desired, such a punishment could be implemented by giving the RL agent negative reward every time step.

Overall, despite differences between artificial and real MWM tasks, we believe that valid comparisons can be made between the two settings, and that our results are slightly closer to humans in virtual MWM tasks than to rodents in real water mazes.
This discussion leads us to make two sets of predictions for humans in virtual MWM settings.
With respect to behavior, we predict that humans in a square maze may develop corner testing strategies.
Specifically, the frequency of this strategy may depend on how salient different available global cues are.
As we saw in Fig \ref{fig:envtype_behaviors}, agents in the 4 Wall Color setting have sufficient global cues to navigate by such that corner testing almost never develops.
We imagine that if cues are small or hard to distinguish from one another, human navigators may similarly come to rely on the shape of the environment, leading to corner test-like strategies, but with plentiful, easy to use cues, corner testing will be less prevalent.

The second set of predictions is with respect to auxiliary tasks.
Auxiliary tasks are quite naturally suited to be implemented with humans, especially categorical tasks, which we observe to be more effective in RL agents.
During an experiment, an experimenter (or the environment) could ask the navigator a question related to an auxiliary task, for example, ``what cardinal direction are you closest to facing'' (Faced Wall), and then provide the correct answer.
Based on our results, we predict that such additional challenges would provide extra learning opportunities, improving navigation abilities over an experiment.

\section{Discussion}

Navigation is a rich domain for reinforcement learning, offering a wide variety of scenarios and environments that can reveal differences in behavior and strategy usage.
The MWM is particularly suitable for studying animals and humans in, as well as being well-suited for simulation as a 2D RL environment.
Simpler 2D environments are less commonly used in deep RL research, but they allow for quick iterative development while still maintaining sufficient complexity for interesting findings.

A key focus of our work is the potential benefit of assigning auxiliary tasks to improve RL effectiveness.
Although it is good to devise general auxiliary tasks that are not environment specific, they are not guaranteed to provide benefit in every situation, such as the Terminal Prediction task, which shows significant improvements in {\it Pommerman} but not in our MWM.
On the other hand, environment-specific auxiliary tasks have potential for enhancing learning.
In particular, we note that easier tasks (i.e., categorical prediction tasks versus numerical prediction ones) often have a greater ability to improve performance.
From a biological perspective, such tasks may be more realistic for real-life agents to perform.
These categorical tasks do not require extremely precise supervision to compute error signals from.
Rather, they only use categorical information, sometimes as simple as binary classification in the case of Left Right tasks.
Animals might even have such information internally available to perform navigation learning with.

One interesting future direction of this research is investigating why auxiliary tasks improve performance while others are not.
In the present work, we looked at this question with a method shown to work in the past \cite{du:2018} - comparing gradient vectors generated by the RL algorithm to those generated by auxiliary tasks.
As we show in Section \ref{sec:grad_cos_sim}, we have not been able to explain differences in task effectiveness based on this measure.
Our current hypothesis, based on empirical observations of our simulations, is that the success of an auxiliary task might be related to how easy the task is for an agent to learn.
In future work, one could approach this hypothesis by attempting to quantify how easy tasks are in terms of how much exposure agents need to learn them.
For example, in the Left Right tasks, we can count the distribution of samples (how often the correct answer is left vs. right), or in the Angle Distance task, we can see how well spread the distribution of correct answers is during training.
This may give a hint to how much learning each task provides, and subsequently how difficult they are to learn.

We also explore methods of measuring representations developed in the activations of RL agent neural networks.
Most real-world scenarios of biological interest are ones where the true state of the environment is unknown, requiring agents to have an internal memory or representation of the environment state.
During regular behavior, some nodes display consistent activation patterns which we can map to places in space or to the direction the agent is facing.
Our representation scores behave in some intuitive ways. 
Agents trained with auxiliary tasks that we would expect to clearly benefit from knowing position or direction tend to earn the highest respective scores. 
Better representation scores also correlate with increased spatial strategy usage.
However, these scores have limitations. 
They are not designed to be sensitive to activation patterns that are periodic in space or direction.
Although our direction representation score has better correlation to direct strategy usage than the spatial representation score, this could be related to the definition or parameter selections of the representation scores.
Correlations between representation scores and performance may also be task dependent.
Having direction representations may be useful in our MWM environment where the goal is fixed, but if the task required exploration within an episode, then spatial representations may become more important.

The ability to observe the development of representations leads to the natural question of whether these nodes can be used for a pre-trained network.
It seems that auxiliary tasks encourage development of representations relatively early in training.
This may explain their negative influence on early performance, as auxiliary tasks can cause the network to rapidly change early on, and policy updates must adjust to the adapting network weights.
After around $10^6$ time steps of experience, however, the representation scores appear to stabilize. 
At this point, the policy can make use of the more useful developed representations, leading to more advanced strategies.
Future experiments could explore freezing early layers of networks that have developed representations and using them as a pre-trained network for a new naive agent. 
One could investigate how pre-existing network weights and representations affect training, and fully decouple representation development from policy optimization to make use of these representations.

{\bf Acknowledgements.} This research was partially funded by the National Science Foundation. We thank Dr. Sarah Creem-Regehr for helpful discussions of this work. The support and resources from the Center for High Performance Computing at the University of Utah are also gratefully acknowledged.

\bibliographystyle{plain}
\bibliography{refs}

\end{document}